\pdfoutput=1
\PassOptionsToPackage{table}{xcolor}
\documentclass[11pt]{article}
\usepackage[final]{ACL2023}
\usepackage{times}
\usepackage{latexsym}
\usepackage[T1]{fontenc}
\usepackage[utf8]{inputenc}
\usepackage{placeins}
\usepackage{microtype}
\usepackage{inconsolata}
\usepackage{amssymb}
\usepackage{multirow}
\usepackage{graphicx}
\usepackage{float}
\usepackage[table]{xcolor}
\definecolor{blue}{HTML}{AADDFF}
\definecolor{green}{HTML}{AAFFAA}
\definecolor{yellow}{HTML}{FFFF88}
\definecolor{red}{HTML}{FF8888}
\definecolor{orange}{HTML}{FFBB55}
\definecolor{purple}{HTML}{AAAAFF}
\definecolor{pink}{HTML}{FFAAFF}
\definecolor{grey}{HTML}{BBBBBB}
\definecolor{color1}{HTML}{00AA00}
\definecolor{color2}{HTML}{AA0000}
\definecolor{color3}{HTML}{0000FF}

\title{Improving Multilingual Capabilities with Cultural and Local Knowledge \\in Large Language Models While Enhancing Native Performance}
\author{
 \textbf{Ram Mohan Rao Kadiyala \textsuperscript{1,3}},
 \textbf{Siddartha Pullakhandam \textsuperscript{2}},\\
 \textbf{Siddhant Gupta         \textsuperscript{3,4}},
 \textbf{Jebish Purbey          \textsuperscript{3}},
 \textbf{Drishti Sharma         \textsuperscript{3}},
 \textbf{Kanwal Mehreen         \textsuperscript{3}},\\
 \textbf{Muhammad Arham         \textsuperscript{3}},
 \textbf{Suman Debnath          \textsuperscript{5}},
 \textbf{Hamza Farooq           \textsuperscript{1,6}}\\
 \\
 \textsuperscript{1}Traversaal.ai, 
 \textsuperscript{2}Vantager, 
 \textsuperscript{3}Cohere for AI Community,\\
 \textsuperscript{4}IIT Roorkee,
 \textsuperscript{5}Amazon, 
 \textsuperscript{6}Stanford University,\\
 \\
 \small{\textbf{Correspondence:} \href{mailto:contact@rkadiyala.com}{ram@traversaal.ai}}\\
 \small{\textbf{Models \& Datasets:} \href{https://huggingface.co/collections/1-800-LLMs/phi-4-hindi-67a75a74e4e1a1b586b4e436}{hf.co/collections/1-800-LLMs/}}
}
\begin{document}
\maketitle
\begin{abstract}
Large Language Models (LLMs) have shown remarkable capabilities, but their development has primarily focused on English and other high-resource languages, leaving many languages underserved. We present our latest Hindi-English bilingual LLM with ~3\% average improvement in benchmark scores over both languages, outperforming models twice its size. Using a curated dataset composed of English and Hindi instruction data of 485K samples, we instruction-tuned models such as Qwen-2.5-14B-Instruct and Phi-4 to improve performance over both English and Hindi. Our experiments, encompassing seven different LLMs of varying parameter sizes and over 140 training attempts with varying English-Hindi training data ratios, demonstrated that it is possible to significantly improve multilingual performance without compromising native performance. Further, our approach avoids resource-intensive techniques like vocabulary expansion or architectural modifications, thus keeping the model size small. Our results indicate that modest fine-tuning with culturally and locally informed data can bridge performance gaps without incurring significant computational overhead. We release our training code, datasets, and models under MIT and Apache licenses to aid further research towards under-represented and low-resource languages.
\end{abstract}
\section{Introduction}
\label{sec:introduction}
The rapid advancement of Large Language Models (LLMs) has led to great advances in various natural language processing tasks. However, the majority of research efforts have disproportionately focused on English and a select few high-resource languages. This disparity leaves a vast number of languages underserved, limiting the global accessibility and applicability of LLM technology. While the lack of readily available data for many languages is a contributing factor, it is not the sole reason. Economic factors and limited access to computational resources also play significant roles in accessibility to the target audience. In this work, we address the gap by developing a bilingual LLM that performs well on English and Hindi tasks. We focused on maintaining relatively smaller model sizes, and rather than resorting to resource-intensive methods such as vocabulary expansion, block expansion, or additional layers, we employ computationally efficient fine-tuning methods such as Supervised Fine-Tuning (SFT) \citep{huggingface2025sfttrainer, vonwerra2022trl} with Low-Rank Adaptation (LoRA) \citep{hu2021lora} through Unsloth \citep{unsloth}. Our primary goal was to boost performance over Hindi tasks while retaining similar performance over English. 

We demonstrate our method by fine-tuning Qwen-2.5-14B-Instruct \citep{qwen2025qwen25technicalreport} and Phi-4 \citep{abdin2024phi4technicalreport} models on a mixed-language dataset. Moreover, our experiments extend to five other LLMs: Gemma 2 9B, Gemma 2 2B \citep{gemmateam2024gemma2improvingopen}, Llama 3.1 8B, Llama 3.1 3B \citep{grattafiori2024llama3herdmodels}, and Qwen 2.5 3B, where over 140 fine-tuning attempts were conducted by varying the distribution ratios of Hindi and English samples of each domain in the training data. These experiments provide insights into how performance changes with varying dataset distributions over each domain. This can help in dataset curation to effectively balance bilingual performance. The promising results suggest that enhancing low-resource language capabilities doesn't necessarily require large-scale architectural changes but can be achieved through targeted, efficient fine-tuning of models with basic capabilities over a language. 
\section{Related Works}
\label{sec:relatedworks}
Prior studies have attempted to address this disparity through various techniques, including vocabulary expansion/modification \citep{tejaswi2024exploring, csaki2023efficiently, shi2024continual, balachandran2023tamil}, modifications in architecture like block expansion and the addition of extra layers to accommodate linguistic diversity \citep{nanda2024llama3}, or continued pre-training followed by instruction tuning again \citep{mahdizadeh-sani-etal-2025-extending, kuulmets-etal-2024-teaching, cui2023efficient, vo2024redwhaleadaptedkoreanllm, luukkonen2023fingpt, kallappa2025krutrim, toraman2024llamaturkadaptingopensourcegenerative}. However, such methods often incur substantial computational costs and lead to an increase in model sizes. 
Prior works also include multilingual LLMs optimized for several languages, including Hindi: Bloom-176B \citep{workshop2023bloom176bparameteropenaccessmultilingual}, Aya-23B \citep{aryabumi2024aya}, Aya-101 \citep{ustun2024aya}, and Aya-expanse \citep{dang2024aya}. Additionally, we also have several other monolingual and bilingual LLMs focused on Hindi: llama-nanda-10B \citep{nanda2024llama3}, Airavata-7B \citep{gala2024airavata}, \citep{BhabhaAI2024}, Aryabhatta-8.5B \citep{GenVRadmin2024Aryabhatta}, Sarvam-2B \citep{sarvamai2024sarvam2b}, Krutrim-2-12B \citep{kallappa2025krutrim}, and Nemotron-mini-Hindi \citep{joshi2024adapting}. The key differences can be seen in \autoref{table:0}.
\begin{table*}
    \centering
    \begin{tabular}{ccccc}
    \noalign{\hrule height 2pt}
    \rowcolor{green} \small{\textbf{Prior Work}} & \small{\textbf{Model}} & \small{\textbf{VRAM Req}} & \small{\textbf{Approach}} & \small{\textbf{Status}}    \\
    \noalign{\hrule height 2pt}
    \small{\citep{GenVRadmin2024Aryabhatta}} & \small{Aryabhatta-8.5B}         & \small{18 GB} & \small{Vocabulary Expansion} & \small{Open-Weights} \\
    \hline
    \small{\citep{gala2024airavata}}         & \small{Airavata-7B}             & \small{14 GB} & \small{LoRA fine-tuning} & \small{Open-Weights} \\
                                                                                                              &&&& \small{Open-Dataset} \\
    \hline
    \small{\citep{sarvamai2024sarvam2b}}     & \small{Sarvam-1-2B}             & \small{6 GB} & \small{From Scratch} & \small{Open-Weights} \\
    \hline
    \small{\citep{kallappa2025krutrim}}      & \small{Krutrim-2-12B-Instruct}  & \small{50 GB} & \small{Vocabulary Expansion} & \small{Open-Weights} \\
    \hline
    \small{\citep{joshi2024adapting}}        & \small{Nemotron-mini-Hindi}     & \small{9 GB} & \small{Continued-Pretraining} & \small{Open-Weights} \\
    \hline
    \small{\citep{nanda2024llama3}}          & \small{Llama-nanda-10B-chat}    & \small{41 GB} & \small{Block Expansion} & \small{Open-Weights} \\
                                                                                     &&& \small{Continued-Pretraining} & \\
    \hline
    \small{\citep{dang2024aya}}              & \small{Aya-Expanse-8B}          & \small{17 GB} & \small{From Scratch} & \small{Open-Weights} \\
    \hline
    \small{\citep{dang2024aya}}              & \small{Aya-Expanse-35B}         & \small{66 GB} & \small{From Scratch} & \small{Open-Weights} \\
    \noalign{\hrule height 2pt}
    \small{Our Work}                         & \small{placeholder-name}        & \small{30 GB} & \small{LoRA with modified} & \small{Open-Dataset} \\
                                            &                                &              & \small{chat template}      & \small{Open-Weights} \\
    \noalign{\hrule height 2pt}
    \end{tabular}
    \caption{Key differences between other works regarding Hindi LLMs}
    \label{table:0}
\end{table*}
\section{Datasets}
\label{sec:datasets}
Despite the existence of datasets to cover several domains for Hindi \citep{khan2024indicllmsuite}, \citep{ramesh2022samanantar}, we decided to experiment primarily with translated/reformatted datasets that do not prohibit usage for research/commercial purposes. This was done so that the same work can be implemented/extended to low-resource languages. Also, fine-tuning on translated data is an efficient way to adapt mPLMs to new languages, leveraging their pre-trained multilingual knowledge. \citep{chen-chen-2024-efficient}. For translation, we used GPT-4o-mini \citep{openai2024gpt4ocard} through Microsoft Azure \footnote{\url{https://azure.microsoft.com/en-us/products/ai-services/openai-service/}} to translate a few datasets and benchmarks from English to Hindi: \nocite{welbl-etal-2017-crowdsourcing}\nocite{lai2017racelargescalereadingcomprehension}\nocite{hendrycks2020measuring}\nocite{cobbe2021training}\nocite{li2023camel}\nocite{clark2019boolq}\nocite{pal2022medmcqa}\nocite{mitra2024orca}\nocite{sakaguchi2021winogrande}\nocite{clark2018think}\nocite{yumetamath}\nocite{zhou2023instructionfollowingevaluationlargelanguage}\nocite{bisk2020piqa} Big-Bench-Hard \citep{suzgun2022challenging}, XNLI \citep{conneau2018xnli}, and XL-Sum \citep{hasan2021xl}. Some of the benchmarks that already have Hindi subsets were used directly: Global MMLU \citep{singh2024global}, IndicXNLI \citep{aggarwal2022indicxnli}. Some of the publicly available datasets containing cultural and localized general knowledge, like Indian legal FAQ \citep{Aditya2411LawIndia2024}, UPSC FAQ \citep{prnv19UPSCFAQ2024}, IndianTAX FAQ \citep{msinankhan1IndiaTaxFAQs2024}, IndianMedicines, IndiaCuisines, and IndiaTravel Guide \citep{cyberblipTravelIndia2024}, were used to generate instruction-response pairs from the tabular format data using GPT-4o-mini as a part of our dataset collection. These were first translated to the other language from the original language and then manually verified by multiple annotators to ensure quality in both languages. We also used a few subsets from the Aya collection \citep{singh-etal-2024-aya}, i.e., the translation, simplification, and summarization subsets. In total the collected dataset had 3.12M samples with a nearly 50:50 ratio of English and Hindi data. Around 90K samples from these cover localized and cultural knowledge. Among the rest, some domains and tasks had a higher proportion in the collection. We used randomly selected subsets from those datasets while maintaining equal language ratios. After filtering the training data, we had around 485K samples, of which 20\% are of localized domain and cultural knowledge, while the rest are of generic tasks like math, MCQs, reasoning, summarization, rephrasing, and translation.
\section{Instruction Data Formatting}
\label{sec:instructdataformatting}
During training we have appended the inputs with different strings based on the task at hand. The details of the appended strings for each task type can be seen in \autoref{table:1}. The underlined portions were replaced with the corresponding texts for each sample. This modification helped in tuning the model to obey instructions well with fewer additional tokens needed for formatting instructions, while not compromising the performance in both languages. The inputs were preprocessed to replace consecutive spaces with a single space, remove leading and trailing spaces, and replace double quotes with single quotes. The same chat templates were used as the original models, with input portions processed into our format.

\begin{table}[!ht]
    \centering
    \begin{tabular}{ll}
    \noalign{\hrule height 2pt}
    \cellcolor{green}\tiny{\textbf{Task}}     & \cellcolor{green}\tiny{\textbf{Input Format}} \\
    \noalign{\hrule height 2pt}
    \tiny{Natural Language Inference}         & \tiny{"\underline{\emph{Text1}} \#\#\# \tiny{\emph{Text2}} \#\#\# \textbf{NLI} \#\#\# :"}\\
    \tiny{Multiple Choice Questions}          & \tiny{"\underline{\emph{Question}} \#\#\# A) \underline{\emph{a}}, B) \underline{\emph{b}},... \#\#\# \textbf{MCQ} \#\#\# :"}\\
    \tiny{Numeric Questions}                  & \tiny{"\underline{\emph{Question}} \#\#\# \textbf{NUMERIC} \#\#\# :"}\\
    \tiny{Boolean Questions}                  & \tiny{"\underline{\emph{Question}} \#\#\# \textbf{BOOLEAN} \#\#\# :"}\\
    \tiny{Questions seeking Long responses}   & \tiny{"\underline{\emph{Question}} \#\#\# \textbf{LONG RESPONSE} \#\#\# :"}\\
    \tiny{Short responses (few words)}        & \tiny{"\underline{\emph{Input}} \#\#\# \textbf{DIRECT RESPONSE} \#\#\# :"}\\
    \tiny{Coding}                             & \tiny{"\underline{\emph{Input}} \#\#\# \textbf{CODE} \#\#\# :"}\\
    \tiny{Text Summarization}                 & \tiny{"\underline{\emph{Input}} \#\#\# \textbf{SUMMARIZE} \#\#\# :"}\\
    \tiny{Paraphrasing/Rephrasing}            & \tiny{"\underline{\emph{Input}} \#\#\# \textbf{PARAPHRASE} \#\#\# :"}\\
    \tiny{Translation to specified language}  & \tiny{"\underline{\emph{Input}} \#\#\# \textbf{TRANSLATION} [\underline{\emph{lang}}] \#\#\# :"}\\
    \tiny{Text Simplification/ELI5}           & \tiny{"\underline{\emph{Input}} \#\#\# \textbf{SIMPLIFY} \#\#\# :"} \\
    \noalign{\hrule height 2pt}    
    \end{tabular}
    \caption{Formats of Input Texts used in training}
    \label{table:1}
\end{table}
\begin{table*}[p]
    \centering
    \begin{tabular}{cc|cc|cc|cc|cc|cc|ccc}
    \noalign{\hrule height 2pt}
    \rowcolor{green}    \textbf{\tiny{Benchmarks}} & \textbf{\tiny{Ratio of}} & \multicolumn{2}{|c|}{\tiny{\textbf{ARC-Challenge}}} & \multicolumn{2}{|c|}{\tiny{\textbf{ARC-Easy}}} & \multicolumn{2}{|c|}{\tiny{\textbf{MMLU}}} & \multicolumn{2}{|c|}{\tiny{\textbf{BoolQ}}} & \multicolumn{2}{|c|}{\tiny{\textbf{Context-MCQ}}} & \multicolumn{3}{|c|}{\tiny{\textbf{Overall Average}}}\\
    \cline{2-15}
    \rowcolor{green}    \textbf{\tiny{Domain data used?}} & \textbf{\tiny{Hindi}} & \textbf{\tiny{En}} & \textbf{\tiny{Hi}} & \textbf{\tiny{En}} & \textbf{\tiny{Hi}} & \textbf{\tiny{En}} & \textbf{\tiny{Hi}} & \textbf{\tiny{En}} & \textbf{\tiny{Hi}} & \textbf{\tiny{En}} & \textbf{\tiny{Hi}} & \textbf{\tiny{En}} & \textbf{\tiny{Hi}} & \textbf{\tiny{Tot}}\\
    \noalign{\hrule height 2pt}
        \tiny{No}  & \tiny{10\%}            & \tiny{90.61} & \tiny{73.21} & \tiny{94.82} & \tiny{80.05} & \tiny{75.74} & \tiny{53.60} & \tiny{84.16} & \tiny{77.24} & \tiny{91.4} & \tiny{79.7}         & \tiny{87.34} & \tiny{72.76} & \tiny{80.05} \\
        \tiny{No}  & \tiny{20\%}            & \tiny{90.53} & \tiny{73.04} & \tiny{94.99} & \tiny{80.68} & \tiny{75.84} & \tiny{53.95} & \tiny{83.30} & \tiny{75.80} & \tiny{90.9} & \tiny{79.0}         & \tiny{87.11} & \tiny{72.49} & \tiny{79.80} \\
        \tiny{No}  & \tiny{30\%}            & \tiny{90.78} & \tiny{73.55} & \tiny{95.16} & \tiny{80.89} & \tiny{75.67} & \tiny{54.00} & \tiny{81.22} & \tiny{74.03} & \tiny{91.2} & \tiny{78.5}         & \tiny{86.80} & \tiny{72.19} & \tiny{79.50} \\
        \tiny{No}  & \tiny{40\%}            & \tiny{91.13} & \tiny{73.29} & \tiny{94.95} & \tiny{80.64} & \tiny{76.09} & \tiny{53.85} & \tiny{84.25} & \tiny{72.29} & \tiny{91.1} & \tiny{78.1}         & \tiny{87.50} & \tiny{71.63} & \tiny{79.57} \\
        \tiny{No}  & \tiny{50\%}            & \tiny{91.30} & \tiny{73.38} & \tiny{94.99} & \tiny{81.19} & \tiny{75.63} & \tiny{54.21} & \tiny{81.53} & \tiny{73.63} & \tiny{91.0} & \tiny{79.0}         & \tiny{86.89} & \tiny{72.28} & \tiny{79.59} \\
        \tiny{No}  & \tiny{60\%}            & \tiny{\textbf{\cellcolor{white}91.55}} & \tiny{75.17} & \tiny{\textbf{\cellcolor{white}95.75}} & \tiny{81.73} & \tiny{75.20} & \tiny{54.29} & \tiny{85.78} & \tiny{75.83} & \tiny{91.7} & \tiny{79.7}         & \tiny{88.00} & \tiny{73.35} & \tiny{80.67} \\
        \tiny{No}  & \tiny{70\%}            & \tiny{91.38} & \tiny{74.91} & \tiny{95.71} & \tiny{82.28} & \tiny{75.52} & \tiny{54.32} & \tiny{85.08} & \tiny{80.82} & \tiny{90.7} & \tiny{79.7}         & \tiny{87.68} & \tiny{74.41} & \tiny{81.04} \\
        \tiny{No}  & \tiny{80\%}            & \tiny{91.13} & \tiny{74.66} & \tiny{94.99} & \tiny{82.37} & \tiny{75.87} & \tiny{54.53} & \tiny{84.19} & \tiny{78.07} & \tiny{91.4} & \tiny{78.8}         & \tiny{87.51} & \tiny{73.68} & \tiny{80.60} \\
        \tiny{No}  & \tiny{90\%}            & \tiny{91.47} & \tiny{75.09} & \tiny{95.50} & \tiny{82.83} & \tiny{75.59} & \tiny{54.69} & \tiny{84.19} & \tiny{79.44} & \tiny{91.2} & \tiny{79.5}         & \tiny{87.59} & \tiny{74.30} & \tiny{80.95} \\
        \tiny{No}  & \tiny{100\%}           & \tiny{91.64} & \tiny{74.83} & \tiny{95.50} & \tiny{\textbf{\cellcolor{white}82.87}} & \tiny{75.69} & \tiny{54.47} & \tiny{85.05} & \tiny{79.72} & \tiny{\textbf{\cellcolor{white}91.6}} & \tiny{\textbf{\cellcolor{white}80.3}}         & \tiny{87.90} & \tiny{74.44} & \tiny{81.17} \\
    \hline
        \tiny{Yes} & \tiny{10\%}            & \tiny{90.96} & \tiny{72.70} & \tiny{94.74} & \tiny{80.26} & \tiny{75.90} & \tiny{53.78} & \tiny{88.47} & \tiny{81.12} & \tiny{90.4} & \tiny{77.3}         & \tiny{88.09} & \tiny{73.03} & \tiny{80.56} \\
        \tiny{Yes} & \tiny{20\%}            & \tiny{90.87} & \tiny{73.29} & \tiny{94.82} & \tiny{81.10} & \tiny{75.89} & \tiny{53.77} & \tiny{88.69} & \tiny{84.27} & \tiny{91.1} & \tiny{78.1}         & \tiny{88.27} & \tiny{74.11} & \tiny{81.19} \\
        \tiny{Yes} & \tiny{30\%}            & \tiny{91.04} & \tiny{73.63} & \tiny{94.91} & \tiny{81.40} & \tiny{75.74} & \tiny{54.24} & \tiny{88.07} & \tiny{81.95} & \tiny{90.8} & \tiny{78.6}         & \tiny{88.11} & \tiny{73.96} & \tiny{81.04} \\
        \tiny{Yes} & \tiny{40\%}            & \tiny{90.78} & \tiny{74.91} & \tiny{94.78} & \tiny{81.65} & \tiny{76.22} & \tiny{54.71} & \tiny{88.78} & \tiny{83.85} & \tiny{90.9} & \tiny{78.8}         & \tiny{88.29} & \tiny{74.78} & \tiny{81.53} \\
        \tiny{Yes} & \tiny{50\%}            & \tiny{91.04} & \tiny{74.74} & \tiny{94.78} & \tiny{81.86} & \tiny{\textbf{\cellcolor{white}76.34}} & \tiny{54.80} & \tiny{88.69} & \tiny{84.61} & \tiny{91.1} & \tiny{78.5}         & \tiny{\textbf{\cellcolor{white}88.39}} & \tiny{74.90} & \tiny{81.64} \\
        \tiny{Yes} & \tiny{60\%}            & \tiny{91.04} & \tiny{75.00} & \tiny{94.87} & \tiny{81.86} & \tiny{75.96} & \tiny{54.76} & \tiny{88.62} & \tiny{84.58} & \tiny{90.9} & \tiny{79.0}         & \tiny{88.27} & \tiny{75.04} & \tiny{81.65} \\
        \tiny{Yes} & \tiny{70\%}            & \tiny{90.87} & \tiny{74.15} & \tiny{94.53} & \tiny{82.11} & \tiny{75.46} & \tiny{54.91} & \tiny{87.86} & \tiny{84.06} & \tiny{91.2} & \tiny{79.7}         & \tiny{87.98} & \tiny{74.98} & \tiny{81.48} \\
        \tiny{Yes} & \tiny{80\%}            & \tiny{90.96} & \tiny{\textbf{\cellcolor{white}76.62}} & \tiny{94.87} & \tiny{82.37} & \tiny{76.04} & \tiny{54.19} & \tiny{\textbf{\cellcolor{white}88.69}} & \tiny{\textbf{\cellcolor{white}84.89}} & \tiny{90.9} & \tiny{78.4}         & \tiny{88.29} & \tiny{75.29} & \tiny{81.79} \\
        \tiny{Yes} & \tiny{90\%}            & \tiny{91.47} & \tiny{75.60} & \tiny{94.74} & \tiny{82.53} & \tiny{75.84} & \tiny{54.77} & \tiny{87.79} & \tiny{84.89} & \tiny{90.8} & \tiny{79.7}         & \tiny{88.15} & \tiny{75.50} & \tiny{81.82} \\
        \tiny{Yes} & \tiny{100\%}           & \tiny{91.21} & \tiny{75.94} & \tiny{94.61} & \tiny{82.70} & \tiny{75.79} & \tiny{\textbf{\cellcolor{white}55.00}} & \tiny{88.29} & \tiny{84.55} & \tiny{91.6} & \tiny{79.7}         & \tiny{88.30} & \tiny{\textbf{\cellcolor{white}75.58}} & \tiny{\textbf{\cellcolor{white}81.94}} \\
    \noalign{\hrule height 2pt}
    \multicolumn{2}{|c|}{\tiny{Original}}   & \tiny{90.87} & \tiny{69.62} & \tiny{95.45} & \tiny{78.49} & \tiny{74.37} & \tiny{52.16} & \tiny{86.09} & \tiny{78.89} & \tiny{91.2} & \tiny{77.4}         & \tiny{87.60} & \tiny{71.31} & \tiny{79.46} \\
    \noalign{\hrule height 2pt}
    \end{tabular}
    \caption{Results (.2f) from each training attempt with 8\% of our training data over Qwen 2.5 14B }
    \label{table:2}
\end{table*}
\begin{table*}[p]
\centering
\begin{tabular}{cc|cc|cc|cc|cc|cc|ccc}
 \noalign{\hrule height 2pt}
 \rowcolor{green}   \textbf{\tiny{Benchmarks}}& \textbf{\tiny{Ratio of}} & \multicolumn{2}{|c|}{\tiny{\textbf{ARC-Challenge}}} & \multicolumn{2}{|c|}{\tiny{\textbf{ARC-Easy}}} & \multicolumn{2}{|c|}{\tiny{\textbf{MMLU}}} & \multicolumn{2}{|c|}{\tiny{\textbf{BoolQ}}} & \multicolumn{2}{|c|}{\tiny{\textbf{Context-MCQ}}} & \multicolumn{3}{|c|}{\tiny{\textbf{Overall Average}}} \\
 \cline{2-15}
 \rowcolor{green}   \textbf{\tiny{Domain data used?}}& \textbf{\tiny{Hindi}} & \textbf{\tiny{En}} & \textbf{\tiny{Hi}} & \textbf{\tiny{En}} & \textbf{\tiny{Hi}} & \textbf{\tiny{En}} & \textbf{\tiny{Hi}} & \textbf{\tiny{En}} & \textbf{\tiny{Hi}} & \textbf{\tiny{En}} & \textbf{\tiny{Hi}} & \textbf{\tiny{En}} & \textbf{\tiny{Hi}} & \textbf{\tiny{Tot}} \\
 \noalign{\hrule height 2pt}
    \tiny{No} & \tiny{10\%} & \tiny{92.24} & \tiny{74.74} & \tiny{97.35} & \tiny{83.67} & \tiny{76.04} & \tiny{50.45} & \tiny{87.52} & \tiny{83.88} & \tiny{86.7} & \tiny{74.7} & \tiny{87.97} & \tiny{73.48} & \tiny{80.72} \\
    \tiny{No} & \tiny{20\%} & \tiny{92.06} & \tiny{75.77} & \tiny{97.39} & \tiny{84.18} & \tiny{76.01} & \tiny{51.61} & \tiny{87.13} & \tiny{83.33} & \tiny{87.0} & \tiny{75.0} & \tiny{87.91} & \tiny{73.97} & \tiny{80.94} \\
    \tiny{No} & \tiny{30\%} & \tiny{92.24} & \tiny{76.54} & \tiny{97.26} & \tiny{84.26} & \tiny{76.02} & \tiny{51.40} & \tiny{87.43} & \tiny{84.22} & \tiny{86.7} & \tiny{75.6} & \tiny{87.93} & \tiny{74.40} & \tiny{81.16} \\
    \tiny{No} & \tiny{40\%} & \tiny{92.15} & \tiny{77.30} & \tiny{97.35} & \tiny{84.97} & \tiny{76.08} & \tiny{51.76} & \tiny{87.16} & \tiny{83.79} & \tiny{\textbf{\cellcolor{white}87.2}} & \tiny{76.1} & \tiny{87.98} & \tiny{74.78} & \tiny{81.38} \\
    \tiny{No} & \tiny{50\%} & \tiny{92.24} & \tiny{82.59} & \tiny{97.43} & \tiny{89.39} & \tiny{76.34} & \tiny{57.41} & \tiny{87.61} & \tiny{85.10} & \tiny{86.6} & \tiny{77.7} & \tiny{88.04} & \tiny{78.43} & \tiny{83.24} \\
    \tiny{No} & \tiny{60\%} & \tiny{92.24} & \tiny{77.39} & \tiny{97.26} & \tiny{84.76} & \tiny{75.82} & \tiny{51.72} & \tiny{87.46} & \tiny{83.91} & \tiny{86.8} & \tiny{75.5} & \tiny{87.91} & \tiny{74.65} & \tiny{81.28} \\
    \tiny{No} & \tiny{70\%} & \tiny{91.98} & \tiny{77.65} & \tiny{97.18} & \tiny{84.89} & \tiny{75.68} & \tiny{51.87} & \tiny{87.49} & \tiny{83.88} & \tiny{86.8} & \tiny{75.8} & \tiny{87.82} & \tiny{74.81} & \tiny{81.32} \\
    \tiny{No} & \tiny{80\%} & \tiny{91.21} & \tiny{77.30} & \tiny{97.31} & \tiny{84.64} & \tiny{75.75} & \tiny{51.59} & \tiny{87.31} & \tiny{84.34} & \tiny{86.2} & \tiny{76} & \tiny{87.55} & \tiny{74.77} & \tiny{81.16} \\
    \tiny{No} & \tiny{90\%} & \tiny{92.32} & \tiny{77.30} & \tiny{97.35} & \tiny{84.51} & \tiny{75.68} & \tiny{50.96} & \tiny{87.58} & \tiny{84.37} & \tiny{86.6} & \tiny{76.1} & \tiny{87.90} & \tiny{74.64} & \tiny{81.27} \\
    \tiny{No} & \tiny{100\%} & \tiny{92.41} & \tiny{78.16} & \tiny{97.39} & \tiny{85.35} & \tiny{75.87} & \tiny{52.12} & \tiny{87.58} & \tiny{83.88} & \tiny{86.1} & \tiny{76.4} & \tiny{87.87} & \tiny{75.18} & \tiny{81.52} \\
  \hline
    \tiny{Yes} & \tiny{10\%} & \tiny{92.15} & \tiny{76.96} & \tiny{97.85} & \tiny{85.31} & \tiny{75.66} & \tiny{50.54} & \tiny{88.53} & \tiny{85.31} & \tiny{86.3} & \tiny{75.0} & \tiny{88.10} & \tiny{74.63} & \tiny{81.36} \\
    \tiny{Yes} & \tiny{20\%} & \tiny{92.49} & \tiny{77.05} & \tiny{97.56} & \tiny{85.69} & \tiny{75.49} & \tiny{50.06} & \tiny{\textbf{\cellcolor{white}88.87}} & \tiny{85.29} & \tiny{86.4} & \tiny{74.5} & \tiny{88.16} & \tiny{74.52} & \tiny{81.34} \\
    \tiny{Yes} & \tiny{30\%} & \tiny{92.49} & \tiny{78.41} & \tiny{97.69} & \tiny{86.95} & \tiny{75.85} & \tiny{51.28} & \tiny{88.35} & \tiny{85.44} & \tiny{86.5} & \tiny{75.4} & \tiny{88.18} & \tiny{75.50} & \tiny{81.84} \\
    \tiny{Yes} & \tiny{40\%} & \tiny{92.66} & \tiny{82.25} & \tiny{97.77} & \tiny{90.36} & \tiny{75.86} & \tiny{56.32} & \tiny{88.65} & \tiny{\textbf{\cellcolor{white}85.92}} & \tiny{86.7} & \tiny{78.3} & \tiny{88.33} & \tiny{\textbf{\cellcolor{white}82.25}} & \tiny{\textbf{\cellcolor{white}83.48}} \\
    \tiny{Yes} & \tiny{50\%} & \tiny{\textbf{\cellcolor{white}93.17}} & \tiny{\textbf{\cellcolor{white}82.93}} & \tiny{\textbf{\cellcolor{white}97.85}} & \tiny{\textbf{\cellcolor{white}91.07}} & \tiny{\textbf{\cellcolor{white}76.52}} & \tiny{\textbf{\cellcolor{white}57.87}} & \tiny{88.31} & \tiny{85.22} & \tiny{87.1} & \tiny{\textbf{\cellcolor{white}78.7}} & \tiny{\textbf{\cellcolor{white}88.59}} & \tiny{79.16} & \tiny{81.88} \\
    \tiny{Yes} & \tiny{60\%} & \tiny{92.49} & \tiny{78.83} & \tiny{97.51} & \tiny{87.07} & \tiny{75.91} & \tiny{52.04} & \tiny{88.07} & \tiny{84.21} & \tiny{86.6} & \tiny{75.9} & \tiny{88.11} & \tiny{75.61} & \tiny{81.86} \\
    \tiny{Yes} & \tiny{70\%} & \tiny{92.40} & \tiny{79.18} & \tiny{97.64} & \tiny{86.70} & \tiny{75.94} & \tiny{51.84} & \tiny{88.31} & \tiny{83.97} & \tiny{86.1} & \tiny{75.8} & \tiny{88.08} & \tiny{75.49} & \tiny{81.79} \\
    \tiny{Yes} & \tiny{80\%} & \tiny{92.66} & \tiny{79.35} & \tiny{97.56} & \tiny{87.75} & \tiny{76.04} & \tiny{52.05} & \tiny{88.13} & \tiny{84.34} & \tiny{85.9} & \tiny{76.6} & \tiny{88.06} & \tiny{76.02} & \tiny{82.04} \\
    \tiny{Yes} & \tiny{90\%} & \tiny{92.58} & \tiny{79.69} & \tiny{97.60} & \tiny{87.96} & \tiny{76.06} & \tiny{52.49} & \tiny{88.23} & \tiny{84.25} & \tiny{86.3} & \tiny{76.4} & \tiny{88.15} & \tiny{76.16} & \tiny{82.16} \\
    \tiny{Yes} & \tiny{100\%} & \tiny{92.49} & \tiny{80.12} & \tiny{97.69} & \tiny{87.58} & \tiny{75.95} & \tiny{52.55} & \tiny{88.32} & \tiny{84.52} & \tiny{86.0} & \tiny{76.2} & \tiny{88.09} & \tiny{76.19} & \tiny{82.14} \\
  \noalign{\hrule height 2pt}
    \multicolumn{2}{|c|}{\tiny{Original}} & \tiny{92.41} & \tiny{79.18} & \tiny{97.31} & \tiny{86.87} & \tiny{74.67} & \tiny{53.24} & \tiny{86.30} & \tiny{82.72} & \tiny{86.3} & \tiny{75.7} & \tiny{87.40} & \tiny{75.54} & \tiny{81.47} \\
  \noalign{\hrule height 2pt}
\end{tabular}
\caption{Results (.2f) from each training attempt with 8\% of our training data over Phi 4 14B }
\label{table:3}
\end{table*}
\section{Initial Evaluation}
\label{sec:initialevaluation}
Before proceeding to train over the full dataset, we have first experimented through several attempts by training on a subset of our data with/without including training data of benchmarks' domains and by varying the ratio of each language in the dataset used. The subsets contain at most 2000 samples from each dataset source for both languages combined. We used normalized next-token log probabilities for MCQs and Boolean benchmarks during the initial evaluation stage to evaluate the models. We then compared how the scores changed with these variations and compared them with the original models to gather insights into optimal final dataset sampling approaches. The results over Qwen-2.5-14B and Phi-4 can be seen below in \autoref{table:2} and \autoref{table:3}, respectively. It can also be seen that, in case of Qwen, the best results were obtained when ratio of Hindi is higher than 50\% but for Phi-4, the results were better with ratio of Hindi less than 50\%.The results for the rest of the models can be found in \autoref{sec:otherresults}.
\section{Dataset Distribution and Ordering}
\label{sec:datasetdistibutionandordering}
The performance of models from initial tests didn't vary significantly with/without being trained on math data. The performance on math subsets of MMLU as well remained similar in both languages with/without being trained on math samples. Since we would be training on a large number of samples, we decided to still use a considerable amount of math samples. A significant performance gap was observed over Boolean benchmarks with a nearly 3\% increase in English and a 5\% increase in Hindi. Hence, we decided to use a slightly higher amount of Boolean questions' samples in the final dataset. The language ratios for each domain in the final dataset were determined based on the initial training data ratios that gave the best results. The samples of the final dataset were sorted over input lengths in ascending order with a certain number of the longest samples placed in the beginning; this approach could improve batch processing efficiency and training stability \citep{wang2024greats}. This number was set equal to the total effective batch size (i.e., the product of batch size and gradient accumulation steps). The samples related to local and cultural knowledge were then placed such that they are evenly spread out in the dataset except for the initial batch. More info on the dataset can be found in \autoref{sec:datasetinfo}. The training methods and details can be found in \autoref{sec:appendix}. 
\section{End Evaluation}
\label{sec:evaluation}
Apart from the benchmarks seen in \autoref{table:2} and \autoref{table:3}, we perform evaluations over additional benchmarks like\nocite{shi2022language}\nocite{conneau2018xnli} like\nocite{shi2022language, conneau2018xnli} like\nocite{shi2022language} MMLU-Pro \citep{wang2024mmlu}, BigBench-Hard \citep{suzgun2022challenging}, MuSR \citep{sprague2024musrtestinglimitschainofthought}, GPQA \citep{rein2023gpqagraduatelevelgoogleproofqa}, and MATH-Hard \citep{hendrycks2021measuringmathematicalproblemsolving}. We used open-llm-leaderboard \footnote{\url{https://huggingface.co/spaces/open-llm-leaderboard/open_llm_leaderboard}} \citep{open-llm-leaderboard-v2} for evaluation over some of the benchmarks through the eval-harness framework \citep{eval-harness}. \autoref{table:7} demonstrates the performance of our models in comparison with the original models over several benchmarks. We did observe variations in the scores from the open-llm-leaderboard and the corresponding benchmark scores, which were self-reported for the original models. We used the scores from the leaderboard for all models over those benchmarks for reproducibility and a fair comparison. The evaluation methods used can be seen in \autoref{table:4}.
\begin{table}[h!t]
    \centering
    \begin{tabular}{ccc}
    \noalign{\hrule height 2pt}
    \rowcolor{green} \textbf{\small{Benchmark}} & \textbf{\small{Eval Criteria}} & \textbf{\small{Eval Framework}} \\
    \noalign{\hrule height 2pt}
        \small{ARC-C}       & \small{0-Shot} & \small{log probabilities} \\
        \small{ARC-E}       & \small{0-Shot} & \small{log probabilities} \\
        \small{BoolQ}       & \small{0-Shot} & \small{log probabilities} \\
        \small{CMCQ}        & \small{0-Shot} & \small{log probabilities} \\
        \small{MMLU}        & \small{0-Shot} & \small{log probabilities} \\
    \hline
        \small{MMLU-Pro}    & \small{5-Shot} & \small{eval-harness} \\
        \small{BBH}         & \small{3-Shot} & \small{eval-harness} \\
        \small{GPQA}        & \small{0-Shot} & \small{eval-harness} \\
        \small{MATH Hard}   & \small{4-Shot} & \small{eval-harness} \\
        \small{MuSR}        & \small{0-Shot} & \small{eval-harness} \\
    \noalign{\hrule height 2pt}
    \end{tabular}
    \caption{Benchmarks used for evaluation and their details}
    \label{table:4}
\end{table}
\section{Generative tasks evaluation}
\label{sec:generative}
\begin{figure}[!h]
    \centering
    \includegraphics[width=1\linewidth]{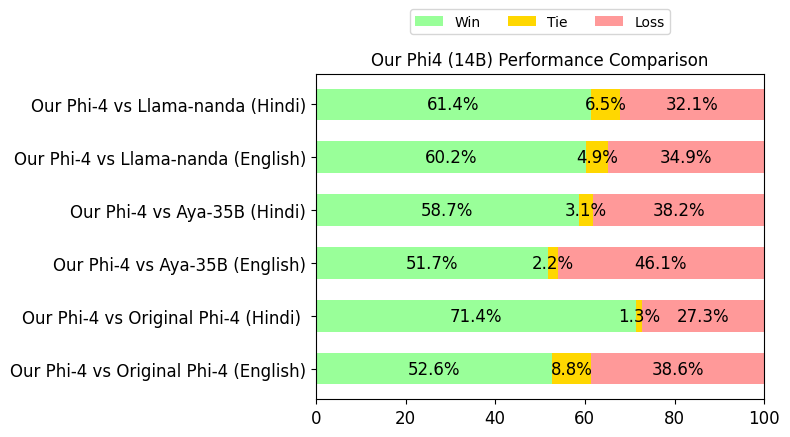}
    \caption{Win Rates with comparable models through human evaluation.}
    \label{figure:0}
\end{figure}

Scarcity of genuine and authentic multilingual benchmarks of a broad range of topics has been a concern for many languages. Prior works in comparison, like \citep{nanda2024llama3}, have not included generative evaluations over either language. while \citep{joshi2024adapting} utilized limited generative benchmarks using LLM-as-a-judge to score the responses, with only the MT-Bench, a translation task, undergoing human evaluation. Further, training on translated data to test over benchmarks translated from English defeats the purpose of building multilingual and multicultural LLMs. \citep{aryabumi2024aya} also utilizes translated benchmarks for multilingual generative task evaluation, with additional human evaluation without topic/domain restriction. We have performed human evaluation in the same way over both languages. These results can be seen in \autoref{figure:0}. 
We performed human evaluations through third-party annotators over both languages over a few of the models that achieved comparably good performance over non-English discriminative tasks. A total of 3217 comparisons were done primarily in Hindi (2097) and the rest in English (1120). For a fair comparison, we utilized the default hyperparameters of each of the models.
\begin{table*}[!ht]
    \centering
    \begin{tabular}{ccccccc|ccccc}
    \noalign{\hrule height 2pt}
    \rowcolor{green}   \tiny{\textbf{Model $\downarrow$}} & \tiny{\textbf{ARC-C}} & \tiny{\textbf{ARC-E}} & \tiny{\textbf{BoolQ}} & \tiny{\textbf{CMCQ}} & \tiny{\textbf{MMLU}} & \tiny{\textbf{Average*}} & \tiny{\textbf{MMLU-Pro}} & \tiny{\textbf{GPQA}} & \tiny{\textbf{MuSR}} & \tiny{\textbf{BBH}} & \tiny{\textbf{MATH}} \\
    \noalign{\hrule height 2pt} 
        \tiny{AryaBhatta-GemmaUltra-8.5B}     & \tiny{22.70} & \tiny{25.04} & \tiny{62.23} & \tiny{22.95} & \tiny{23.70}   & \tiny{31.32}  & \tiny{22.66} & \tiny{25.34} & \tiny{42.72} & \tiny{41.12} & \tiny{ 2.95}\\
        \tiny{Airavata-7B}                    & \tiny{25.09} & \tiny{30.47} & \tiny{62.17} & \tiny{25.31} & \tiny{33.20}   & \tiny{35.25}  & \tiny{16.35} & \tiny{27.43} & \tiny{37.57} & \tiny{36.00} & \tiny{13.60}\\  
        \tiny{sarvam-1-2B}                    & \tiny{30.03} & \tiny{33.25} & \tiny{62.17} & \tiny{42.80} & \tiny{27.90}   & \tiny{39.23}  & \tiny{-}     & \tiny{-}     & \tiny{-}     & \tiny{-}     & \tiny{-}\\
        \tiny{Nemotron-4-Mini-Hindi-Instruct} & \tiny{55.80} & \tiny{71.63} & \tiny{62.11} & \tiny{68.10} & \tiny{43.20}   & \tiny{60.17}  & \tiny{25.95} & \tiny{30.87} & \tiny{41.53} & \tiny{40.11} & \tiny{ 2.04}\\
        \tiny{Llama-3-Nanda-10B-Chat}         & \tiny{65.36} & \tiny{80.64} & \tiny{82.29} & \tiny{67.60} & \tiny{50.61}   & \tiny{69.30}  & \tiny{31.57}     & \tiny{30.11}     & \tiny{43.52}     & \tiny{49.38}     & \tiny{ 5.59}\\
        \tiny{Krutrim-2-12b-instruct}         & \tiny{67.32} & \tiny{81.10} & \tiny{84.74} & \tiny{76.30} & \tiny{56.10}   & \tiny{73.11}  & \tiny{-}     & \tiny{-}     & \tiny{-}     & \tiny{-}     & \tiny{-}\\
        \tiny{aya-expanse-8b}                 & \tiny{74.06} & \tiny{87.08} & \tiny{86.45} & \tiny{83.30} & \tiny{56.89}   & \tiny{77.56}  & \tiny{30.04} & \tiny{30.29} & \tiny{37.17} & \tiny{49.42} & \tiny{ 7.02}\\
        \tiny{aya-expanse-32B}                & \tiny{85.41} & \tiny{\textbf{\underline{95.08}}} & \tiny{\textbf{\underline{90.43}}} & \tiny{\underline{89.80}} & \tiny{69.71}   & \tiny{86.08}  & \tiny{41.30} & \tiny{32.55} & \tiny{38.62} & \tiny{56.29} & \tiny{13.37}\\
    \noalign{\hrule height 2pt}
        \tiny{Our Qwen Model (14B)}           & \tiny{\textbf{\underline{90.61}}} & \tiny{\underline{94.82}} & \tiny{{88.53}} & \tiny{\textbf{\underline{90.70}}} & \tiny{\underline{75.00}}   & \tiny{\underline{87.93}}  & \tiny{\textbf{\underline{52.63}}} & \tiny{\underline{36.24}} & \tiny{\underline{44.84}} & \tiny{{64.97}} & \tiny{\textbf{\underline{25.08}}}\\
        \tiny{Our Qwen Model (T)}             & \tiny{90.47}                      & \tiny{\underline{94.82}} & \tiny{\underline{88.59}} & \tiny{{89.69}} & \tiny{{74.81}} & \tiny{{87.68}}     & \tiny{{\underline{52.58}}} & \tiny{{36.09}} & \tiny{{44.77}} & \tiny{{\underline{65.04}}} & \tiny{{\underline{24.32}}} \\
        \tiny{Our Phi Model (14B)}            & \tiny{\textbf{\underline{97.39}}} & \tiny{92.24} & \tiny{87.65} & \tiny{87.40} & \tiny{\textbf{\underline{75.59}}}   & \tiny{\textbf{\underline{88.05}}}  & \tiny{52.39} & \tiny{\textbf{\underline{39.77}}} & \tiny{\textbf{\underline{49.07}}} & \tiny{\textbf{\underline{66.97}}} & \tiny{{23.11}}\\
    \noalign{\hrule height 2pt}
    \end{tabular}
    \caption{Metrics (.2f) of our and other LLMs over several \textbf{English} benchmarks.}
    \caption*{*Averages for English were calculated using just the first 5 benchmarks for similar comparison with Hindi}
    \label{table:5}
 \end{table*}
\begin{table*}[!h]
    \centering
    \begin{tabular}{ccccccc}
    \noalign{\hrule height 2pt}
    \rowcolor{green}\tiny{\textbf{Model $\downarrow$}} & \tiny{\textbf{ARC-C}} & \tiny{\textbf{ARC-E}} & \tiny{\textbf{BoolQ}} & \tiny{\textbf{CMCQ}} & \tiny{\textbf{MMLU}} & \tiny{\textbf{Average}} \\
    \noalign{\hrule height 2pt}
        \tiny{AryaBhatta-GemmaUltra-8.5B}           & \tiny{22.70} & \tiny{25.08} & \tiny{62.17} & \tiny{22.95} & \tiny{23.80}   & \tiny{31.34} \\
        \tiny{Airavata-7B}                          & \tiny{22.87} & \tiny{25.13} & \tiny{62.17} & \tiny{23.28} & \tiny{33.20}   & \tiny{33.33} \\
        \tiny{sarvam-1-2B}                          & \tiny{32.76} & \tiny{35.06} & \tiny{62.16} & \tiny{47.10} & \tiny{24.22}   & \tiny{40.26} \\
        \tiny{Llama-3-Nanda-10B-Chat}               & \tiny{45.99} & \tiny{60.56} & \tiny{71.96} & \tiny{54.70} & \tiny{36.35}   & \tiny{53.91} \\
        \tiny{Nemotron-4-Mini-Hindi-4B-Instruct}    & \tiny{50.68} & \tiny{63.72} & \tiny{68.74} & \tiny{51.30} & \tiny{37.18}   & \tiny{54.32} \\
        \tiny{Krutrim-2-12b-instruct}               & \tiny{56.83} & \tiny{70.66} & \tiny{78.86} & \tiny{64.10} & \tiny{46.51}   & \tiny{63.39} \\
        \tiny{aya-expanse-8b}                       & \tiny{57.42} & \tiny{72.90} & \tiny{80.42} & \tiny{69.00} & \tiny{43.39}   & \tiny{64.63} \\
        \tiny{aya-expanse-32B}                      & \tiny{73.29} & \tiny{\underline{85.48}} & \tiny{\textbf{\underline{87.73}}} & \tiny{\textbf{\underline{79.70}}} & \tiny{\textbf{\underline{56.96}}}   & \tiny{\underline{76.63}} \\
    \noalign{\hrule height 2pt}
        \tiny{Our Qwen Model (14B)}     & \tiny{74.06} & \tiny{81.23} & \tiny{84.07} & \tiny{78.20} & \tiny{53.85}   & \tiny{74.82} \\
        \tiny{Ouw Qwen Model (T)}       & \tiny{74.84} & \tiny{81.38} & \tiny{84.97} & \tiny{75.38} & \tiny{52.92}   & \tiny{73.91} \\
        \tiny{Our Phi Model (14B)}      & \tiny{\textbf{\underline{81.74}}} & \tiny{\textbf{\underline{89.06}}} & \tiny{\underline{86.02}} & \tiny{\underline{78.70}} & \tiny{\underline{56.39}}   & \tiny{\textbf{\underline{78.38}}} \\
    \noalign{\hrule height 2pt}
    \end{tabular}
    \caption{Metrics (.2f) of our and other LLMs over several \textbf{Hindi} benchmarks}
    \label{table:6}
\end{table*}
\section{Comparisons}
\label{sec:comparisions}
For additional comparisons, we compare the performance of our models with other Hindi bilingual LLMs and other open-source LLMs that are optimized for Hindi. Due to the large variations in the number of parameters of our models and other comparable models, we compare average benchmark performance versus the model size in terms of VRAM requirement. The comparisons over English and Hindi benchmarks alongside our Qwen and Phi models can be seen in \autoref{table:5} and \autoref{table:6}. Over the benchmarks of higher difficulty, our models have consistently outperformed models over twice their size, as seen in \autoref{table:5}. The comparison also includes a version of Qwen that was trained on purely translated data, unlike the other two, where a translated dataset is used in cases of missing data. This was done by translating the original dataset to English and back-translating back to Hindi. \footnote{Qwen trained purely on translated data produced nearly identical results on both languages' benchmarks compared to the main model trained on a mix of real and translated data.}
\begin{table*}[!h]
    \centering
    \begin{tabular}{cccccccc}
    \noalign{\hrule height 2pt}
    \rowcolor{green}   \textbf{\tiny{Benchmark}}  & \textbf{\tiny{Lang}} & \textbf{\tiny{Qwen-2.5-}} & \textbf{\tiny{Our Qwen}} & \textbf{\tiny{Change}} & \textbf{\tiny{Phi-4}} & \textbf{\tiny{Our Phi-4}} & \textbf{\tiny{Change}} \\
    \rowcolor{green}    & & \textbf{\tiny{14B-Instruct}} & & & & & \\
    \noalign{\hrule height 2pt}
       \multirow{2}{*}{\tiny{ARC-Easy}}       & \tiny{En} & \tiny{95.45} & \tiny{94.82} & {\color{color2}\tiny{$\blacktriangledown$ 0.63}}    & \tiny{97.31} & \tiny{97.39} & {\color{color1}\tiny{$\blacktriangle$ 0.08}} \\
       \cline{2-8}                                         
                                         & \tiny{Hi} & \tiny{78.49} & \tiny{81.23} & {\color{color1}\tiny{$\blacktriangle$ 2.74}}        & \tiny{86.87} & \tiny{89.06} & {\color{color1}\tiny{$\blacktriangle$ 2.19}} \\
       \hline                                        
       \multirow{2}{*}{\tiny{ARC-Challenge}}  & \tiny{En} & \tiny{90.87} & \tiny{90.61} & {\color{color2}\tiny{$\blacktriangledown$ 0.26}}    & \tiny{92.41} & \tiny{92.24} & {\color{color2}\tiny{$\blacktriangledown$ 0.17}} \\
       \cline{2-8}
                                         & \tiny{Hi} & \tiny{69.62} & \tiny{74.06} & {\color{color1}\tiny{$\blacktriangle$ 4.44}}        & \tiny{79.18} & \tiny{81.74} & {\color{color1}\tiny{$\blacktriangle$ 2.56}} \\
       \hline
       \multirow{2}{*}{\tiny{BoolQ}}          & \tiny{En} & \tiny{86.09} & \tiny{88.53} & {\color{color1}\tiny{$\blacktriangle$ 2.44}}        & \tiny{86.30} & \tiny{87.65} & {\color{color1}\tiny{$\blacktriangle$ 1.35}} \\
       \cline{2-8}
                                         & \tiny{Hi} & \tiny{78.89} & \tiny{84.07} & {\color{color1}\tiny{$\blacktriangle$ 5.18}}        & \tiny{82.72} & \tiny{86.02} & {\color{color1}\tiny{$\blacktriangle$ 3.30}} \\
       \hline
       \multirow{2}{*}{\tiny{Context-MCQ}}    & \tiny{En} & \tiny{91.20} & \tiny{90.70} & {\color{color2}\tiny{$\blacktriangledown$ 0.50}}    & \tiny{86.30} & \tiny{87.40} & {\color{color1}\tiny{$\blacktriangle$ 1.10}} \\
       \cline{2-8}       
                                         & \tiny{Hi} & \tiny{77.40} & \tiny{78.20} & {\color{color1}\tiny{$\blacktriangle$ 0.80}}        & \tiny{75.70} & \tiny{78.70} & {\color{color1}\tiny{$\blacktriangle$ 3.00}} \\
       \hline
       \multirow{2}{*}{\tiny{MMLU}}           & \tiny{En} & \tiny{74.37} & \tiny{75.00} & {\color{color1}\tiny{$\blacktriangle$ 0.63}}        & \tiny{74.67} & \tiny{75.59} & {\color{color1}\tiny{$\blacktriangle$ 0.92}} \\
       \cline{2-8}           
                                         & \tiny{Hi} & \tiny{52.16} & \tiny{53.85} & {\color{color1}\tiny{$\blacktriangle$ 1.69}}        & \tiny{53.24} & \tiny{56.39} & {\color{color1}\tiny{$\blacktriangle$ 3.15}} \\
    \noalign{\hrule height 2pt}
     \multirow{2}{*}{\textbf{\tiny{Average}}} &\textbf{\tiny{En}} &\textbf{\tiny{87.60}} &\textbf{\tiny{87.93}} &\textbf{\color{color1}\tiny{$\blacktriangle$ 0.33}} & \textbf{\tiny{87.40}} & \textbf{\tiny{88.05}} & \textbf{\color{color1}\tiny{$\blacktriangle$ 0.65}} \\
                                       &\textbf{\tiny{Hi}} &\textbf{\tiny{71.31}} &\textbf{\tiny{74.82}} &\textbf{\color{color1}\tiny{$\blacktriangle$ 3.51}} & \textbf{\tiny{75.54}} & \textbf{\tiny{78.38}} & \textbf{\color{color1}\tiny{$\blacktriangle$ 2.84}} \\
    \noalign{\hrule height 2pt}
                            & \textbf{\tiny{Overall}} & \textbf{\tiny{79.46}} & \textbf{\tiny{81.38}} & \textbf{\color{color1}\tiny{$\blacktriangle$ 1.92}} & \textbf{\tiny{81.47}} & \textbf{\tiny{83.22}} & \textbf{\color{color1}\tiny{$\blacktriangle$ 1.75}} \\
    \noalign{\hrule height 2pt}   
    \end{tabular}
    \caption{Performance of our models compared to originals over each benchmark : evals through log likelihoods}
    \label{table:7}
\end{table*}
\begin{table*}[!h]
    \centering
    \begin{tabular}{cccccccc}
    \noalign{\hrule height 2pt}
    \rowcolor{green}   \textbf{{Benchmark}}  & \textbf{{Lang}} & \textbf{{Qwen-2.5-}} & \textbf{{Our Qwen}} & \textbf{{Change}} & \textbf{{Phi-4}} & \textbf{{Our Phi-4}} & \textbf{{Change}} \\
    \rowcolor{green}    & & \textbf{{14B-Instruct}} & & & & & \\
    \noalign{\hrule height 2pt}
       {MMLU-Pro}                        & En & 49.04 & 52.63 & {\color{color1}$\blacktriangle$ 3.59}        & 53.78 & 52.39 & {\color{color2}$\blacktriangledown$ 1.39} \\
       \hline
       {MATH hard}                       & En & 00.00 & 25.08 & {\color{color3}$\blacktriangle$ N/A}         & 12.31 & 23.11 & {\color{color1}$\blacktriangle$ 10.80} \\
       \hline
       {GPQA}                            & En & 32.21 & 36.24 & {\color{color1}$\blacktriangle$ 4.03}        & 33.72 & 39.77 & {\color{color1}$\blacktriangle$ 6.05} \\
       \hline
       {MuSR}                            & En & 40.87 & 44.84 & {\color{color1}$\blacktriangle$ 3.97}        & 41.01 & 49.07 & {\color{color1}$\blacktriangle$ 8.06} \\
       \hline
       {BigBench-Hard}                   & En & 63.74 & 64.97 & {\color{color1}$\blacktriangle$ 1.23}        & 68.60 & 66.97 & {\color{color2}$\blacktriangledown$ 1.63} \\                 
    \noalign{\hrule height 2pt}  
       \textbf{Average}                  &    & \textbf{37.17} & \textbf{44.75} & \textbf{\color{color1}$\blacktriangle$ 7.58} & \textbf{41.88} & \textbf{46.26} & \textbf{\color{color1}$\blacktriangle$ 4.38} \\
    \noalign{\hrule height 2pt}
    \end{tabular}
    \caption{Performance of our models compared to originals over each benchmark : evals through eval-harness}
    \label{table:8}
\end{table*}
\subsection{Domain wise Performance change}
\label{subsec:domainwiseperformance}
The performance of our models compared to the original versions over MMLU-pro can be seen in \autoref{table:9}. The type of questions the models faced through MMLU-Pro may be of the same domain but were of different subdomains and task types compared to those in our datasets. For example, the CS benchmarks' questions were MCQs about various topics in computer science, while our training data over CS was from MBPP \citep{austin2021program} alone, which consists of a text input and a Python code as an output. Further, the only source of training data we used for economics consists of tax filing FAQs in the Indian context and primarily in Hindi. Hence, such domains' data usage was mentioned as N/A. The domains that had a performance boost in our models without being in training data had questions of the form of fill-mask or text completion, which were similar to the training data from Winogrande-XL \citep{sakaguchi2021winogrande} and PIQA \citep{bisk2020piqa} spanning several domains. 
\begin{figure}[!h]
    \centering
    \includegraphics[width=1\linewidth]{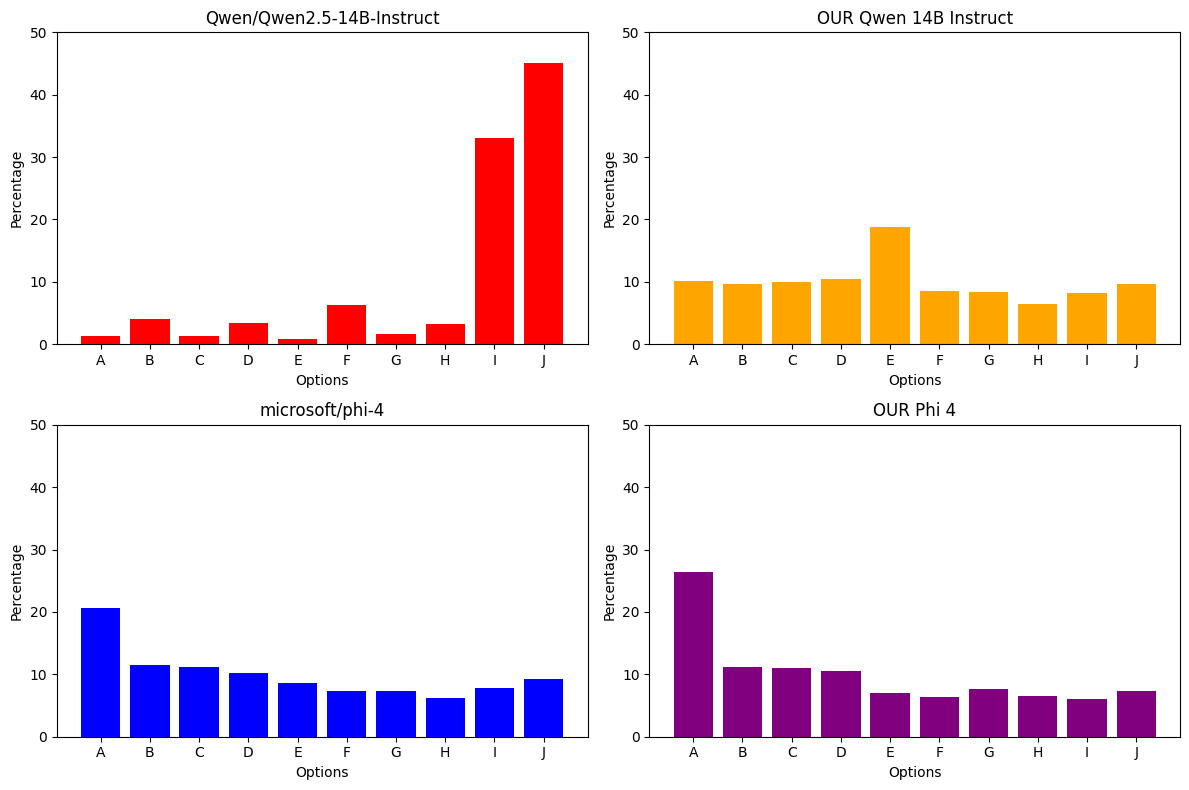}
    \caption{Distribution of each model's choices over MMLU-Pro}
    \label{figure:1}
\end{figure}
\begin{table*}[!h]
    \centering
    \begin{tabular}{cccccccc}
    \noalign{\hrule height 2pt}
    \rowcolor{green}    \textbf{\tiny{Model}} $\rightarrow$ & \multicolumn{2}{|c|}{\textbf{\tiny{Qwen-2.5-14B}}} & \textbf{\tiny{Change}} & \multicolumn{2}{|c|}{\textbf{\tiny{Phi-4}}} & \textbf{\tiny{Change}} & \textbf{\tiny{Training}}  \\
    \noalign{\hrule height 2pt}    
    \rowcolor{green}    \textbf{\tiny{Domain}} $\downarrow$ & \textbf{\tiny{Original}} & \textbf{\tiny{Ours}}           &                 & \textbf{\tiny{Original}} & \textbf{\tiny{Ours}}    &                 & \textbf{\tiny{Data Used}} \\
    \noalign{\hrule height 2pt}
        \tiny{Health}              & \tiny{60.39} & \tiny{65.65} & \color{color1}\tiny{$\blacktriangle$ 5.26}       & \tiny{65.40} & \tiny{65.40} & \color{color1}\tiny{$\blacktriangle$ 0.00}       & \tiny{Yes}   \\
        \tiny{Biology}             & \tiny{76.15} & \tiny{79.36} & \color{color1}\tiny{$\blacktriangle$ 3.21}       & \tiny{80.89} & \tiny{81.03} & \color{color1}\tiny{$\blacktriangle$ 0.14}       & \tiny{Yes}   \\
        \tiny{Engineering}         & \tiny{38.08} & \tiny{46.85} & \color{color1}\tiny{$\blacktriangle$ 8.77}       & \tiny{47.06} & \tiny{44.17} & \color{color2}\tiny{$\blacktriangledown$ 2.89}   & \tiny{Yes}   \\
        \tiny{Math}                & \tiny{39.53} & \tiny{44.78} & \color{color1}\tiny{$\blacktriangle$ 5.25}       & \tiny{41.01} & \tiny{38.79} & \color{color2}\tiny{$\blacktriangledown$ 2.22}   & \tiny{Yes}   \\
        \tiny{Physics}             & \tiny{39.80} & \tiny{41.96} & \color{color1}\tiny{$\blacktriangle$ 2.16}       & \tiny{42.80} & \tiny{39.11} & \color{color2}\tiny{$\blacktriangledown$ 3.69}   & \tiny{Yes}   \\
        \tiny{Chemistry}           & \tiny{35.78} & \tiny{38.25} & \color{color1}\tiny{$\blacktriangle$ 2.47}       & \tiny{36.75} & \tiny{35.69} & \color{color2}\tiny{$\blacktriangledown$ 1.06}   & \tiny{Yes}   \\
        \tiny{Law}                 & \tiny{37.78} & \tiny{41.42} & \color{color1}\tiny{$\blacktriangle$ 3.64}       & \tiny{48.14} & \tiny{47.14} & \color{color2}\tiny{$\blacktriangledown$ 1.00}   & \tiny{Yes}   \\
        \tiny{Philosophy}          & \tiny{53.51} & \tiny{57.92} & \color{color1}\tiny{$\blacktriangle$ 4.41}       & \tiny{62.32} & \tiny{59.72} & \color{color2}\tiny{$\blacktriangledown$ 2.60}   & \tiny{N/A}    \\
        \tiny{Psychology}          & \tiny{70.05} & \tiny{73.81} & \color{color1}\tiny{$\blacktriangle$ 3.76}       & \tiny{76.32} & \tiny{76.82} & \color{color1}\tiny{$\blacktriangle$ 0.50}       & \tiny{N/A}    \\
        \tiny{Business}            & \tiny{37.90} & \tiny{45.63} & \color{color1}\tiny{$\blacktriangle$ 7.73}       & \tiny{40.94} & \tiny{38.91} & \color{color2}\tiny{$\blacktriangledown$ 2.03}   & \tiny{N/A}    \\
        \tiny{CS}                  & \tiny{50.73} & \tiny{53.17} & \color{color1}\tiny{$\blacktriangle$ 2.44}       & \tiny{60.00} & \tiny{58.78} & \color{color2}\tiny{$\blacktriangledown$ 1.22}   & \tiny{N/A}    \\
        \tiny{Economics}           & \tiny{66.71} & \tiny{66.47} & \color{color2}\tiny{$\blacktriangledown$ 0.24}   & \tiny{68.84} & \tiny{69.08} & \color{color1}\tiny{$\blacktriangle$ 0.26}       & \tiny{No}    \\
        \tiny{History}             & \tiny{58.01} & \tiny{57.74} & \color{color2}\tiny{$\blacktriangledown$ 0.27}   & \tiny{63.78} & \tiny{62.73} & \color{color2}\tiny{$\blacktriangledown$ 1.05}   & \tiny{No}    \\
        \tiny{Other}               & \tiny{54.44} & \tiny{53.68} & \color{color2}\tiny{$\blacktriangledown$ 0.76}   & \tiny{57.47} & \tiny{56.71} & \color{color2}\tiny{$\blacktriangledown$ 0.76}   & \tiny{No}    \\
    \noalign{\hrule height 2pt}
    \end{tabular}
    \caption{Domain wise performance changes over MMLU-Pro (English) with our models}
    \label{table:9}
\end{table*}
\subsection{Model biases over choices}
\label{subsec:modelbiasoverchoices}
The observations from domain-wise performance changes by Phi and Qwen were significantly different. The domains that were well represented in our training data had a significant boost in both languages of MMLU. Despite training on MCQs, which consist of 2-4 options, similar results of improvement were seen over MMLU-Pro, which has up to 10 options. On the other hand, Phi-4 had a higher performance boost over MMLU, which has the same number of options as the samples in the training data, but the performance over MMLU-Pro dropped irrespective of domain. The distribution of choices made by each of our LLMs and the corresponding original implementation can be seen in \autoref{figure:1}. The instruction tuning dataset we used had an equal distribution of each of the choices among MCQ samples. The original Qwen model overwhelmingly chose from the final two options, while our model was able to generalize well despite not being trained on MCQs with 10 choices. On the other hand, the original phi-4 was able to perform better than its counterpart, but despite being fine-tuned with equal distribution of choices, the model displayed an inclination towards the first choice among the list of options. The extent of this bias varied between each domain significantly. More on this can be seen in \autoref{sec:modelchoices}. As our Phi model was fine-tuned from the original models' instruct variants, the biases were assumed to have been carried forward. Our models were able to respond well with fewer biases in choices over the domains whose samples are present in large quantities in our training data. To further look into this, we tried to fine-tune the base variant of qwen-2.5-14B rather than the instruct model to see the choice distribution over MMLU-Pro. While most of our dataset's samples of MCQs had 4-5 samples, it was reflected in the choices made as seen in \autoref{figure:1111}, which demonstrates the issue within the original model similar to previous works demonstrating sensitivity on models' sensitivity to order of choices \citep{pezeshkpour2024large}. But a well-balanced instruction-tuning dataset can minimize this issue or an evaluation independent of the order of choices \citep{zheng2023large}. A slight tilt from left to right in \autoref{figure:1} and \autoref{figure:1111} can be expected, as not all questions are accompanied by 10 options, with a considerable amount having less. 
\section{Conclusion}
\label{sec:conclusion}
We demonstrate that enhancing low-resource language capabilities in LLMs is possible through targeted fine-tuning rather than complex architectural changes. Our work shows that a 12-15B parameter LLM provides an effective balance between performance and accessibility, requiring just 30GB RAM. The performance analysis reveals that our Phi-4 model excels in general-purpose tasks, while the Qwen model shows stronger adaptation to specific domains, as evidenced by the domain-wise performance changes in \autoref{table:9}. Our approach of using primarily translated datasets, except for culturally specific knowledge, makes this method readily adaptable to other low-resource languages. To further push the research in low-resource languages, we release our training code, datasets, and models under commercially permissible licenses.
\subsection{Scalability to other languages}
\label{subsec:scalabilitytoother}
As not every language has readily available datasets of even a few domains, we took an approach of using just translated datasets for all domains other than those used for localized and cultural knowledge addition. This would enable reusing the approach to build bilingual LLMs optimized for other languages as long as a proficient LLM supports the language to translate the texts fluently. Given that the performance of the models trained on a mix of real and translated data as well as just translated data are nearly identical as seen in \autoref{table:5} and \autoref{table:6}. This technique could be scaled to other languages.
\subsection{Model Efficiency}
\label{subsec:modelefficiency}
\begin{figure}[!h]
    \centering
    \includegraphics[width=0.9\linewidth]{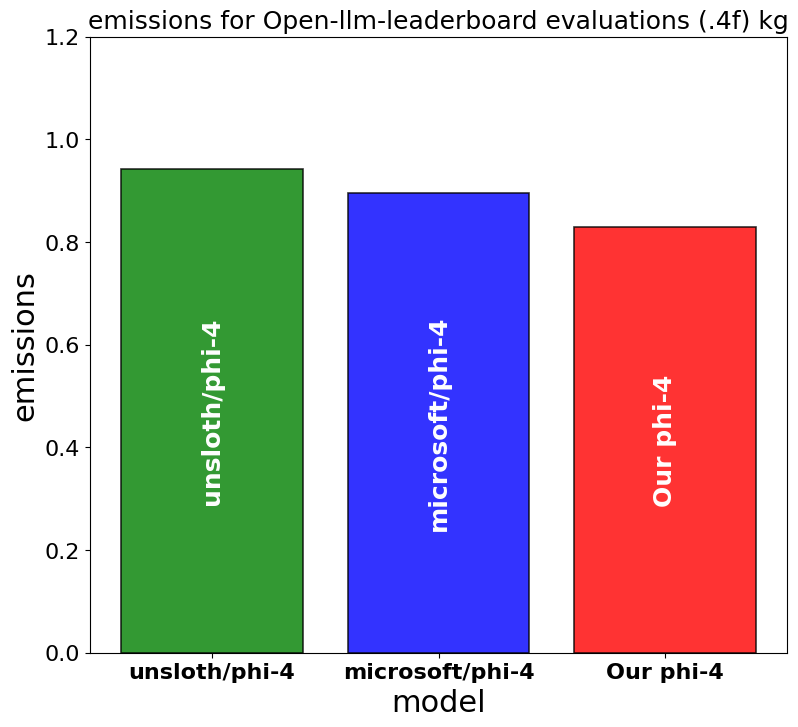}
    \caption{Emissions : open-llm-leaderboard evaluation}
    \label{figure:5}
\end{figure}

Unsloth's version of phi-4 \citep{unsloth-phi4} with Llama architecture led to an improved performance but increased emissions. Our model resulted in lesser emissions during evaluation over the open-llm-leaderboard while improving the model's performance. A comparison of our model to the original and unsloth's phi-4 can be seen in \autoref{figure:5}.
\section*{Limitations}
\label{sec:limitations}
Our models, although demonstrating robust performance across multiple benchmarks, may produce inaccurate, incomplete, or irrelevant outputs due to knowledge cutoffs in its training data. The models although working well directly with the original chat template are better optimized for our prompt formats. The approach presented has been tested in several attempts with Hindi, we believe a similar boost can be obtained over other languages as well, but has not been tested yet.
\bibliography{anthology,custom}
\bibliographystyle{acl_natbib}
\appendix
\section{Model Replication}
\label{sec:appendix}
The hyper-parameters used for training can be seen below in \autoref{table:10}. The initial training attempts using a portion of the data (i.e 8\% samples) were done on various different devices, All experiments combined including evals and training consumed an equivalent of 642 Hours on H200 SXM. 
\begin{figure}[!h]
    \centering
    \includegraphics[width=0.8\linewidth]{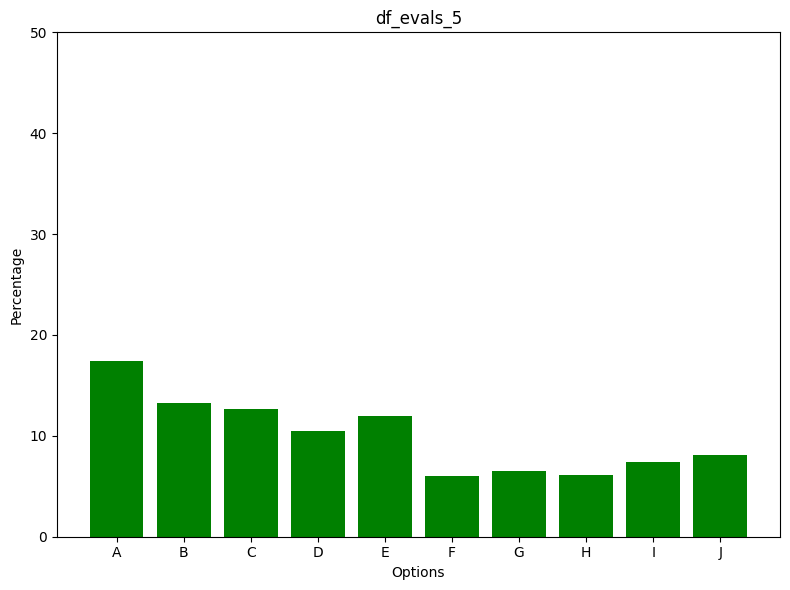}
    \caption{Distribution of response choices of our model training from Qwen-base variant over MMLU-Pro}
    \label{figure:1111}
\end{figure}
\section{License}
\label{sec:license}
Our Qwen and Phi models are available through the same licenses as the models we used as a base, i.e., Apache-2.0 and MIT, respectively. The models can be accessed here. \footnote{Our Phi-4 model :\url{https://huggingface.co/}}. The training datasets are publicly available here. \footnote{Datasets: \url{https://huggingface.co/}}. Most datasets used for training have a copyleft license, with the rest having no license specified and being publicly available on Hugging Face.
\begin{table}[!h]
    \centering
    \begin{tabular}{ccc}
    \noalign{\hrule height 2pt}
    \rowcolor{green}    \multicolumn{2}{c}{\textbf{Hyperparameter}} & \textbf{Value} \\
    \noalign{\hrule height 2pt}
        Seed                & Row Shuffling         & 1024 \\
                            & Dataset Sampling      & 1024 \\
                            & Training              & 1024 \\
                            & Random State          & 1024 \\
    \hline                        
        Epochs              &                       & 1 \\
    \hline
        Total Batch Size    &                       & 600 \\
                            & Batch Size            & 40 \\
                            & \small{Gradient Accumulation}       & 15 \\
    \hline
        Learning Rate       &                       & 2e-5 \\
    \hline
        Weight Decay        &                       & 1e-2 \\
    \hline
        Warmup Steps        &                       & 0 \\
    \noalign{\hrule height 2pt}
    \end{tabular}
    \caption{Training hyper-parameters used}
    \label{table:10}
\end{table}
\begin{table*}[!h]
    \centering
    \begin{tabular}{ccrrrc}
    \noalign{\hrule height 2pt}
    \rowcolor{green}    \textbf{Domain} & \textbf{Dataset} & \textbf{Total}   & \textbf{Used}    & \textbf{Hindi} & \textbf{Original} \\
    \rowcolor{green}                    &                  & \textbf{Samples} & \textbf{Samples} & \textbf{Ratio} & \textbf{Source} \\
    \noalign{\hrule height 2pt}
        Legal FAQ         & India Law       & 51,210  & 51,210    & N/A  &  \citep{Aditya2411LawIndia2024} \\
        Cooking Recipes   & India Recipe    & 13,742  & 13,742    &   *  &  ** \\
        Travel FAQ        & India Travel    & 2,000   & 2,000     & N/A  &  \citep{cyberblipTravelIndia2024} \\
        Tax FAQ           & India TAX       & 2,235   & 2,235     & N/A  &  \citep{msinankhan1IndiaTaxFAQs2024} \\
        General Knowledge & India UPSC      & 620     & 620       & N/A  &  \citep{prnv19UPSCFAQ2024} \\
    \hline
        General           & BoolQ           & 18,799  & 18,799    & N/A  &  \citep{clark2019boolq} \\
        General           & Context MCQs    & 18,505  & 18,505    & N/A  &  \citep{lai2017racelargescalereadingcomprehension} \\
                          &                 &         &           &      &  \citep{welbl2017crowdsourcingmultiplechoicescience} \\
        General           & ARC challenge   & 2,835   & 2,835     & N/A  &  \citep{clark2018think} \\
        General           & ARC Easy        & 5,637   & 5,637     & N/A  &  \citep{clark2018think} \\
        General           & Winogrande XL   & 82,973  & 10,000    &  85  &  \citep{sakaguchi2021winogrande} \\
    \hline
        Biology           & Camel Biology   & 39,990  & 39,990    & N/A  &  \citep{li2023camel} \\
        Biology           & Bio Instruct    & 49,956  & 49,956    & N/A  &  \citep{10.1093/jamia/ocae122} \\
        Coding            & MBPP            & 928     & 928       & N/A  &  \citep{austin2021program} \\
        Chemistry         & Camel Chemistry & 39,975  & 39,975    & N/A  &  \citep{li2023camel} \\
        NLI               & XNLI/IndicXNLI  & 395,192 & 20,000    &  80  &  \citep{conneau2018xnli} \\
                          &                 &         &           &      &  \citep{aggarwal2022indicxnli} \\
        Math              & MATH QA         & 68,583  & 10,000    &  50  &  \citep{amini-etal-2019-mathqa} \\
        Math              & Math Hard       & 4,593   & 4,593     & N/A  &  \citep{hendrycks2021measuringmathematicalproblemsolving} \\
        Math              & Math Easy       & 14,953  & 14,953    & N/A  &  \citep{hendrycks2021measuringmathematicalproblemsolving} \\
        Math              & GSM8K           & 14,937  & 14,973    & N/A  &  \citep{cobbe2021training} \\
        Math              & Camel Math      & 99,626  & 10,000    &  50  &  \citep{li2023camel} \\
        Math              & META Math       & 199,782 & 20,000    &  80  &  \citep{yumetamath} \\
        Math              & Orca Math       & 399,847 & 10,000    &  50  &  \citep{mitra2024orca} \\
        Medical           & MedMCQA         & 372,779 & 20,000    &  70  &  \citep{pal2022medmcqa} \\
        Paraphrasing      & Aya Paraphrase  & 1,001   & 1,001     & N/A  &  \citep{singh-etal-2024-aya} \\
        Physics           & Camel Physics   & 39,995  & 39,995    & N/A  &  \citep{li2023camel} \\
        Reasoning         & PIQA            & 35,396  & 35,396    & N/A  &  \citep{bisk2020piqa} \\
        Reasoning         & SIQA            & 65,630  & 20,000    &  80  &  \citep{sap-etal-2019-social} \\
        Simplification    & Aya Simplify    & 994,944 & 10,000    &  60  &  \citep{singh-etal-2024-aya} \\
        Summarization     & XLSum           & 79,625  & 10,000    &  50  &  \citep{hasan2021xl} \\
        Translation       & Aya Translate   & 1,156   & 1,156     & N/A  &  \citep{singh-etal-2024-aya} \\
    \noalign{\hrule height 2pt}     
                    &      & \textbf{3,117,450} & \textbf{485,469} &     &  \\     
    \noalign{\hrule height 2pt}     
    \end{tabular}
    \caption{Sources of our training dataset's samples and their distributions}
    \caption*{* indicates that the original dataset had a language mix of English and Hindi. Among the rest, initial sample counts were 50:50 for each language and were later individually sampled based on the ratios mentioned for each dataset.}
    \caption*{** The dataset at the time of data collection was publicly available on hf without a restrictive license, but is currently made private.}
    \label{table:11}
\end{table*}

The initially collected dataset sources, sample sizes and the later used sample counts can be seen in \autoref{table:11} along with the ratios of each language. The sampling within each dataset is done at random using the seed specified in \autoref{table:10}. The samples were sorted in ascending order based on input size and the longest 600 samples in terms of input token count were added in the beginning of the training data.
\section{Datasets and Benchmarks Info}
\label{sec:datasetinfo}
The benchmarks used can be seen in \autoref{table:12} along their features like domain, original source, total number of samples, number of samples used and the ratio of Hindi samples among those used.
\begin{table*}
    \centering
    \begin{tabular}{ccc}
    \noalign{\hrule height 2pt}
     \rowcolor{green}   \textbf{Benchmark} & \textbf{Source} \\
    \noalign{\hrule height 2pt}
        ARC Easy      &  \citep{clark2018think} \\
        ARC Challenge &  \citep{clark2018think} \\
        Context MCQs  &  \citep{lai2017racelargescalereadingcomprehension}, \citep{welbl2017crowdsourcingmultiplechoicescience} \\
        BoolQ         &  \citep{clark2019boolq} \\
        MMLU          &  \citep{hendrycks2020measuring}, \citep{singh2024global} \\
    \hline
        MMLU-Pro      &  \citep{wang2024mmlu} \\
        MATH-HARD     &  \citep{hendrycks2021measuringmathematicalproblemsolving} \\
        GPQA          &  \citep{rein2023gpqagraduatelevelgoogleproofqa} \\
        MuSR          &  \citep{sprague2024musrtestinglimitschainofthought} \\
        Bigbench-Hard &  \citep{suzgun2022challenging} \\
    \noalign{\hrule height 2pt}     
    \end{tabular}
    \caption{Benchmarks used and their corresponding sources}
    \label{table:12}
\end{table*}
\section{Results from other attempts}
\label{sec:otherresults}
The results from other attempts with a smaller sized LLMs can be seen in Llama-3.1-8B: \autoref{table:101}, Llama-3.2-3B: \autoref{table:102}, Gemma-2-9B: \autoref{table:103}, Gemma-2-2B: \autoref{table:104}, Qwen-2.5-3B: \autoref{table:105}.
\begin{table*}[!ht]
\centering
\begin{tabular}{cc|cc|cc|cc|cc|cc|ccc}
\noalign{\hrule height 2pt}
\rowcolor{green} \textbf{\tiny{Benchmarks}}& \textbf{\tiny{Ratio of}} & \multicolumn{2}{|c|}{\tiny{\textbf{ARC-Challenge}}} & \multicolumn{2}{|c|}{\tiny{\textbf{ARC-Easy}}} & \multicolumn{2}{|c|}{\tiny{\textbf{MMLU}}} & \multicolumn{2}{|c|}{\tiny{\textbf{BoolQ}}} & \multicolumn{2}{|c|}{\tiny{\textbf{Context-MCQ}}} & \multicolumn{3}{|c|}{\tiny{\textbf{Overall Average}}} \\
     \cline{2-15}
 \rowcolor{green} \textbf{\tiny{Data used?}}& \textbf{\tiny{Hindi}} & \textbf{\tiny{En}} & \textbf{\tiny{Hi}} & \textbf{\tiny{En}} & \textbf{\tiny{Hi}} & \textbf{\tiny{En}} & \textbf{\tiny{Hi}} & \textbf{\tiny{En}} & \textbf{\tiny{Hi}} & \textbf{\tiny{En}} & \textbf{\tiny{Hi}} & \textbf{\tiny{En}} & \textbf{\tiny{Hi}} & \textbf{\tiny{Tot}} \\
\noalign{\hrule height 2pt}
\tiny{No} & \tiny{10\%} & \tiny{78.07} & \tiny{39.51} & \tiny{88.97} & \tiny{47.98} & \tiny{59.42} & \tiny{35.44} & \tiny{62.26} & \tiny{62.25} & \tiny{82.0} & \tiny{56.4} & \tiny{74.14} & \tiny{48.31} & \tiny{61.23} \\
\tiny{No} & \tiny{20\%} & \tiny{77.65} & \tiny{40.19} & \tiny{88.72} & \tiny{50.00} & \tiny{59.92} & \tiny{34.63} & \tiny{62.35} & \tiny{62.28} & \tiny{75.9} & \tiny{53.2} & \tiny{72.91} & \tiny{48.06} & \tiny{60.48} \\
\tiny{No} & \tiny{30\%} & \tiny{77.65} & \tiny{39.51} & \tiny{88.51} & \tiny{49.79} & \tiny{59.33} & \tiny{34.76} & \tiny{62.32} & \tiny{62.16} & \tiny{76.9} & \tiny{55.5} & \tiny{72.94} & \tiny{48.34} & \tiny{60.64} \\
\tiny{No} & \tiny{40\%} & \tiny{77.56} & \tiny{40.44} & \tiny{88.59} & \tiny{50.63} & \tiny{59.92} & \tiny{34.38} & \tiny{62.39} & \tiny{63.35} & \tiny{76.1} & \tiny{52.5} & \tiny{72.91} & \tiny{48.04} & \tiny{60.48} \\
\tiny{No} & \tiny{50\%} & \tiny{78.16} & \tiny{41.89} & \tiny{88.72} & \tiny{50.55} & \tiny{60.97} & \tiny{35.23} & \tiny{62.35} & \tiny{62.31} & \tiny{77.5} & \tiny{54.2} & \tiny{73.54} & \tiny{48.83} & \tiny{61.18} \\
\tiny{No} & \tiny{60\%} & \tiny{78.50} & \tiny{41.81} & \tiny{88.72} & \tiny{50.46} & \tiny{61.00} & \tiny{35.40} & \tiny{62.35} & \tiny{62.31} & \tiny{78.2} & \tiny{54.7} & \tiny{73.75} & \tiny{48.93} & \tiny{61.34} \\
\tiny{No} & \tiny{70\%} & \tiny{78.33} & \tiny{42.06} & \tiny{88.89} & \tiny{50.46} & \tiny{60.85} & \tiny{35.37} & \tiny{62.35} & \tiny{62.31} & \tiny{78.1} & \tiny{54.9} & \tiny{73.70} & \tiny{49.02} & \tiny{61.36} \\
\tiny{No} & \tiny{80\%} & \tiny{78.24} & \tiny{42.32} & \tiny{88.59} & \tiny{50.55} & \tiny{60.86} & \tiny{35.36} & \tiny{62.35} & \tiny{62.31} & \tiny{78.1} & \tiny{55.3} & \tiny{73.62} & \tiny{49.16} & \tiny{61.39} \\
\tiny{No} & \tiny{90\%} & \tiny{76.79} & \tiny{39.76} & \tiny{88.34} & \tiny{45.92} & \tiny{57.91} & \tiny{32.35} & \tiny{62.23} & \tiny{62.19} & \tiny{77.9} & \tiny{50.6} & \tiny{72.63} & \tiny{46.16} & \tiny{59.39} \\
\tiny{No} & \tiny{100\%} & \tiny{75.77} & \tiny{38.91} & \tiny{87.88} & \tiny{45.54} & \tiny{57.76} & \tiny{31.98} & \tiny{62.26} & \tiny{62.19} & \tiny{76.7} & \tiny{50.8} & \tiny{72.07} & \tiny{45.88} & \tiny{58.97} \\
\noalign{\hrule height 2pt}
\tiny{Yes} & \tiny{10\%} & \tiny{78.50} & \tiny{42.32} & \tiny{89.86} & \tiny{50.93} & \tiny{60.03} & \tiny{35.39} & \tiny{71.25} & \tiny{62.74} & \tiny{80.6} & \tiny{56.3} & \tiny{76.04} & \tiny{49.53} & \tiny{62.79} \\
\tiny{Yes} & \tiny{20\%} & \tiny{77.99} & \tiny{39.93} & \tiny{88.80} & \tiny{50.25} & \tiny{59.74} & \tiny{34.51} & \tiny{62.54} & \tiny{62.07} & \tiny{74.5} & \tiny{53.2} & \tiny{72.71} & \tiny{47.99} & \tiny{60.35} \\
\tiny{Yes} & \tiny{30\%} & \tiny{77.82} & \tiny{40.53} & \tiny{88.76} & \tiny{50.42} & \tiny{59.47} & \tiny{34.57} & \tiny{62.75} & \tiny{62.19} & \tiny{74.0} & \tiny{50.9} & \tiny{72.56} & \tiny{47.72} & \tiny{60.14} \\
\tiny{Yes} & \tiny{40\%} & \tiny{77.82} & \tiny{40.53} & \tiny{88.64} & \tiny{50.38} & \tiny{59.67} & \tiny{34.09} & \tiny{62.72} & \tiny{62.22} & \tiny{71.3} & \tiny{49.3} & \tiny{72.03} & \tiny{47.30} & \tiny{59.67} \\
\tiny{Yes} & \tiny{50\%} & \tiny{78.16} & \tiny{41.13} & \tiny{88.59} & \tiny{51.18} & \tiny{60.72} & \tiny{34.95} & \tiny{62.66} & \tiny{62.28} & \tiny{75.2} & \tiny{52.3} & \tiny{73.06} & \tiny{48.36} & \tiny{60.71} \\
\tiny{Yes} & \tiny{60\%} & \tiny{78.50} & \tiny{41.47} & \tiny{88.72} & \tiny{50.42} & \tiny{60.68} & \tiny{35.17} & \tiny{62.45} & \tiny{62.34} & \tiny{76.3} & \tiny{53.1} & \tiny{73.33} & \tiny{48.50} & \tiny{60.91} \\
\tiny{Yes} & \tiny{70\%} & \tiny{78.50} & \tiny{42.06} & \tiny{88.68} & \tiny{50.51} & \tiny{60.71} & \tiny{35.12} & \tiny{62.45} & \tiny{62.37} & \tiny{76.2} & \tiny{53.5} & \tiny{73.30} & \tiny{48.71} & \tiny{61.01} \\
\tiny{Yes} & \tiny{80\%} & \tiny{78.58} & \tiny{42.24} & \tiny{88.72} & \tiny{50.51} & \tiny{60.76} & \tiny{35.24} & \tiny{62.42} & \tiny{62.37} & \tiny{76.6} & \tiny{53.6} & \tiny{73.41} & \tiny{48.79} & \tiny{61.10} \\
\tiny{Yes} & \tiny{90\%} & \tiny{77.22} & \tiny{42.15} & \tiny{88.85} & \tiny{49.87} & \tiny{57.39} & \tiny{30.28} & \tiny{64.86} & \tiny{64.03} & \tiny{69.0} & \tiny{43.7} & \tiny{71.46} & \tiny{46.00} & \tiny{58.73} \\
\tiny{Yes} & \tiny{100\%} & \tiny{75.77} & \tiny{38.91} & \tiny{87.88} & \tiny{45.54} & \tiny{57.76} & \tiny{31.98} & \tiny{63.79} & \tiny{62.80} & \tiny{72.1} & \tiny{43.7} & \tiny{71.46} & \tiny{44.58} & \tiny{58.02} \\
\noalign{\hrule height 2pt}
\multicolumn{2}{|c|}{\tiny{Original}} & \tiny{77.73} & \tiny{41.21} & \tiny{88.26} & \tiny{49.20} & \tiny{60.25} & \tiny{34.26} & \tiny{62.20} & \tiny{62.25} & \tiny{76.3} & \tiny{52.7} & \tiny{72.94} & \tiny{47.92} & \tiny{60.43} \\
\noalign{\hrule height 2pt}
\end{tabular}
\caption{Results (.2f) from each training attempt with 5\% of our training data over Qwen2.5-3B-Instruct}
 \label{table:105}
\end{table*}
\begin{table*}[!h]
  \centering
  \begin{tabular}{cc|cc|cc|cc|cc|cc|ccc}
  \noalign{\hrule height 2pt}
  \rowcolor{green}  \textbf{\tiny{Benchmarks}}     & \textbf{\tiny{Ratio of}} & \multicolumn{2}{|c|}{\tiny{\textbf{ARC-Challenge}}} & \multicolumn{2}{|c|}{\tiny{\textbf{ARC-Easy}}} & \multicolumn{2}{|c|}{\tiny{\textbf{MMLU}}} & \multicolumn{2}{|c|}{\tiny{\textbf{BoolQ}}} & \multicolumn{2}{|c|}{\tiny{\textbf{Context-MCQ}}} & \multicolumn{3}{|c|}{\tiny{\textbf{Overall Average}}} \\
  \cline{2-15} 
  \rowcolor{green}  \textbf{\tiny{Data used?}}     & \textbf{\tiny{Hindi}} & \textbf{\tiny{En}} & \textbf{\tiny{Hi}} & \textbf{\tiny{En}} & \textbf{\tiny{Hi}} & \textbf{\tiny{En}} & \textbf{\tiny{Hi}} & \textbf{\tiny{En}} & \textbf{\tiny{Hi}} & \textbf{\tiny{En}} & \textbf{\tiny{Hi}} & \textbf{\tiny{En}} & \textbf{\tiny{Hi}} & \textbf{\tiny{Tot}} \\
  \noalign{\hrule height 2pt}
    \tiny{No} & \tiny{10\%} & \tiny{73.89} & \tiny{61.06} & \tiny{85.94} & \tiny{66.66} & \tiny{62.30} & \tiny{42.11} & \tiny{64.13} & \tiny{61.06} & \tiny{82.8} & \tiny{64.4} & \tiny{73.81} & \tiny{57.52} & \tiny{65.67} \\
    \tiny{No} & \tiny{20\%} & \tiny{75.43} & \tiny{55.72} & \tiny{87.37} & \tiny{69.40} & \tiny{63.09} & \tiny{42.95} & \tiny{63.94} & \tiny{61.49} & \tiny{83.2} & \tiny{65.3} & \tiny{74.60} & \tiny{58.97} & \tiny{66.78} \\
    \tiny{No} & \tiny{30\%} & \tiny{75.40} & \tiny{55.97} & \tiny{87.04} & \tiny{69.95} & \tiny{62.98} & \tiny{43.03} & \tiny{62.69} & \tiny{59.90} & \tiny{83.2} & \tiny{65.8} & \tiny{74.26} & \tiny{58.93} & \tiny{66.60} \\
    \tiny{No} & \tiny{40\%} & \tiny{73.63} & \tiny{54.86} & \tiny{86.66} & \tiny{68.56} & \tiny{62.34} & \tiny{42.25} & \tiny{63.91} & \tiny{61.76} & \tiny{82.2} & \tiny{65.2} & \tiny{73.74} & \tiny{58.52} & \tiny{66.13} \\
    \tiny{No} & \tiny{50\%} & \tiny{74.23} & \tiny{55.89} & \tiny{86.66} & \tiny{70.12} & \tiny{62.60} & \tiny{42.35} & \tiny{64.80} & \tiny{61.79} & \tiny{82.4} & \tiny{65.0} & \tiny{74.13} & \tiny{59.02} & \tiny{66.58} \\
    \tiny{No} & \tiny{60\%} & \tiny{72.70} & \tiny{54.86} & \tiny{84.81} & \tiny{67.97} & \tiny{60.65} & \tiny{42.06} & \tiny{64.46} & \tiny{60.97} & \tiny{82.1} & \tiny{65.2} & \tiny{72.94} & \tiny{58.21} & \tiny{65.58} \\
    \tiny{No} & \tiny{70\%} & \tiny{75.26} & \tiny{56.23} & \tiny{88.80} & \tiny{69.82} & \tiny{62.53} & \tiny{42.27} & \tiny{65.72} & \tiny{60.14} & \tiny{82.2} & \tiny{64.9} & \tiny{74.90} & \tiny{58.67} & \tiny{66.79} \\
    \tiny{No} & \tiny{80\%} & \tiny{74.23} & \tiny{54.69} & \tiny{86.24} & \tiny{68.10} & \tiny{62.18} & \tiny{42.62} & \tiny{64.53} & \tiny{61.27} & \tiny{81.5} & \tiny{64.9} & \tiny{73.73} & \tiny{58.31} & \tiny{66.02} \\
    \tiny{No} & \tiny{90\%} & \tiny{73.81} & \tiny{54.95} & \tiny{85.90} & \tiny{67.89} & \tiny{61.81} & \tiny{42.33} & \tiny{63.88} & \tiny{61.39} & \tiny{81.3} & \tiny{63.5} & \tiny{73.34} & \tiny{58.01} & \tiny{65.68} \\
    \tiny{No} & \tiny{100\%} & \tiny{73.81} & \tiny{55.03} & \tiny{86.07} & \tiny{68.64} & \tiny{61.57} & \tiny{42.30} & \tiny{63.88} & \tiny{57.48} & \tiny{80.8} & \tiny{64.3} & \tiny{73.22} & \tiny{57.55} & \tiny{65.38} \\
  \noalign{\hrule height 2pt}
    \tiny{Yes} & \tiny{10\%} & \tiny{79.27} & \tiny{59.13} & \tiny{91.50} & \tiny{75.59} & \tiny{63.91} & \tiny{42.49} & \tiny{83.98} & \tiny{74.49} & \tiny{83.5} & \tiny{66.0} & \tiny{80.43} & \tiny{63.54} & \tiny{71.98} \\
    \tiny{Yes} & \tiny{20\%} & \tiny{79.35} & \tiny{58.79} & \tiny{91.41} & \tiny{76.47} & \tiny{64.01} & \tiny{43.65} & \tiny{85.96} & \tiny{79.66} & \tiny{84.5} & \tiny{66.6} & \tiny{81.05} & \tiny{65.03} & \tiny{73.04} \\
    \tiny{Yes} & \tiny{30\%} & \tiny{79.01} & \tiny{61.69} & \tiny{92.47} & \tiny{76.43} & \tiny{64.04} & \tiny{43.17} & \tiny{84.95} & \tiny{77.82} & \tiny{83.4} & \tiny{66.8} & \tiny{80.77} & \tiny{65.18} & \tiny{72.98} \\
    \tiny{Yes} & \tiny{40\%} & \tiny{79.18} & \tiny{61.35} & \tiny{91.62} & \tiny{76.68} & \tiny{63.62} & \tiny{43.27} & \tiny{84.98} & \tiny{74.79} & \tiny{83.7} & \tiny{65.6} & \tiny{80.62} & \tiny{64.34} & \tiny{72.48} \\
    \tiny{Yes} & \tiny{50\%} & \tiny{78.92} & \tiny{60.92} & \tiny{91.67} & \tiny{76.18} & \tiny{62.95} & \tiny{43.15} & \tiny{85.26} & \tiny{78.19} & \tiny{83.8} & \tiny{67.5} & \tiny{80.52} & \tiny{65.19} & \tiny{72.85} \\
    \tiny{Yes} & \tiny{60\%} & \tiny{77.39} & \tiny{60.07} & \tiny{92.00} & \tiny{75.97} & \tiny{63.44} & \tiny{43.43} & \tiny{85.02} & \tiny{78.37} & \tiny{82.2} & \tiny{66.5} & \tiny{80.01} & \tiny{64.87} & \tiny{72.44} \\
    \tiny{Yes} & \tiny{70\%} & \tiny{78.33} & \tiny{61.35} & \tiny{91.71} & \tiny{76.09} & \tiny{63.67} & \tiny{43.41} & \tiny{83.36} & \tiny{75.28} & \tiny{82.7} & \tiny{66.0} & \tiny{79.95} & \tiny{64.45} & \tiny{72.20} \\
    \tiny{Yes} & \tiny{80\%} & \tiny{76.79} & \tiny{58.79} & \tiny{89.73} & \tiny{75.42} & \tiny{62.84} & \tiny{42.91} & \tiny{83.27} & \tiny{74.27} & \tiny{82.2} & \tiny{66.4} & \tiny{78.97} & \tiny{63.56} & \tiny{71.26} \\
    \tiny{Yes} & \tiny{90\%} & \tiny{76.88} & \tiny{59.81} & \tiny{90.40} & \tiny{75.00} & \tiny{62.69} & \tiny{43.06} & \tiny{83.03} & \tiny{73.97} & \tiny{82.0} & \tiny{65.7} & \tiny{79.00} & \tiny{63.51} & \tiny{71.25} \\
    \tiny{Yes} & \tiny{100\%} & \tiny{76.54} & \tiny{59.81} & \tiny{89.73} & \tiny{75.72} & \tiny{62.54} & \tiny{43.70} & \tiny{82.35} & \tiny{77.00} & \tiny{81.2} & \tiny{67.5} & \tiny{78.47} & \tiny{64.74} & \tiny{71.61} \\
  \noalign{\hrule height 2pt}
    \multicolumn{2}{|c|}{\tiny{Original}} & \tiny{75.34} & \tiny{53.92} & \tiny{84.76} & \tiny{65.78} & \tiny{61.69} & \tiny{43.32} & \tiny{65.17} & \tiny{62.16} & \tiny{78.4} & \tiny{67.1} & \tiny{73.07} & \tiny{58.45} & \tiny{65.76} \\
  \noalign{\hrule height 2pt}
    \end{tabular}
    \caption{Results (.2f) from each training attempt with 5\% of our training data over LLama 3.1 8B }
    \label{table:101}
\end{table*}
\begin{table*}[!h]
\centering
\begin{tabular}{cc|cc|cc|cc|cc|cc|ccc}
 \noalign{\hrule height 2pt}
 \rowcolor{green}   \textbf{\tiny{Benchmarks}}& \textbf{\tiny{Ratio of}} & \multicolumn{2}{|c|}{\tiny{\textbf{ARC-Challenge}}} & \multicolumn{2}{|c|}{\tiny{\textbf{ARC-Easy}}} & \multicolumn{2}{|c|}{\tiny{\textbf{MMLU}}} & \multicolumn{2}{|c|}{\tiny{\textbf{BoolQ}}} & \multicolumn{2}{|c|}{\tiny{\textbf{Context-MCQ}}} & \multicolumn{3}{|c|}{\tiny{\textbf{Overall Average}}} \\
\cline{2-15}
 \rowcolor{green}   \textbf{\tiny{Data used?}}& \textbf{\tiny{Hindi}} & \textbf{\tiny{En}} & \textbf{\tiny{Hi}} & \textbf{\tiny{En}} & \textbf{\tiny{Hi}} & \textbf{\tiny{En}} & \textbf{\tiny{Hi}} & \textbf{\tiny{En}} & \textbf{\tiny{Hi}} & \textbf{\tiny{En}} & \textbf{\tiny{Hi}} & \textbf{\tiny{En}} & \textbf{\tiny{Hi}} & \textbf{\tiny{Tot}} \\
\noalign{\hrule height 2pt}
    \tiny{No} & \tiny{10\%} & \tiny{60.83} & \tiny{41.97} & \tiny{75.71} & \tiny{55.47} & \tiny{51.60} & \tiny{33.69} & \tiny{65.44} & \tiny{62.71} & \tiny{68.6} & \tiny{49.1} & \tiny{64.44} & \tiny{48.59} & \tiny{56.51} \\
    \tiny{No} & \tiny{20\%} & \tiny{60.75} & \tiny{43.60} & \tiny{76.85} & \tiny{55.80} & \tiny{52.79} & \tiny{33.86} & \tiny{65.01} & \tiny{62.55} & \tiny{69.2} & \tiny{51.1} & \tiny{64.92} & \tiny{49.38} & \tiny{57.15} \\
    \tiny{No} & \tiny{30\%} & \tiny{60.66} & \tiny{42.32} & \tiny{76.26} & \tiny{55.13} & \tiny{53.28} & \tiny{33.84} & \tiny{64.64} & \tiny{62.19} & \tiny{68.4} & \tiny{51.0} & \tiny{64.65} & \tiny{48.89} & \tiny{56.77} \\
    \tiny{No} & \tiny{40\%} & \tiny{60.49} & \tiny{41.97} & \tiny{75.46} & \tiny{55.13} & \tiny{52.28} & \tiny{33.67} & \tiny{64.46} & \tiny{62.61} & \tiny{69.7} & \tiny{50.9} & \tiny{64.48} & \tiny{48.86} & \tiny{56.67} \\
    \tiny{No} & \tiny{50\%} & \tiny{60.41} & \tiny{44.28} & \tiny{76.09} & \tiny{55.51} & \tiny{51.71} & \tiny{31.63} & \tiny{65.20} & \tiny{62.77} & \tiny{68.0} & \tiny{52.3} & \tiny{64.28} & \tiny{49.30} & \tiny{56.79} \\
    \tiny{No} & \tiny{60\%} & \tiny{60.49} & \tiny{45.56} & \tiny{76.34} & \tiny{56.43} & \tiny{51.24} & \tiny{32.36} & \tiny{65.29} & \tiny{62.98} & \tiny{68.7} & \tiny{51.8} & \tiny{64.41} & \tiny{49.82} & \tiny{57.12} \\
    \tiny{No} & \tiny{70\%} & \tiny{62.20} & \tiny{45.64} & \tiny{77.31} & \tiny{57.23} & \tiny{52.50} & \tiny{32.01} & \tiny{64.98} & \tiny{62.49} & \tiny{68.9} & \tiny{51.5} & \tiny{65.18} & \tiny{49.78} & \tiny{57.48} \\
    \tiny{No} & \tiny{80\%} & \tiny{61.94} & \tiny{44.88} & \tiny{76.85} & \tiny{56.18} & \tiny{52.48} & \tiny{33.06} & \tiny{65.56} & \tiny{61.76} & \tiny{70.4} & \tiny{53.7} & \tiny{61.94} & \tiny{49.91} & \tiny{57.68} \\
    \tiny{No} & \tiny{90\%} & \tiny{63.31} & \tiny{46.84} & \tiny{77.99} & \tiny{58.21} & \tiny{49.12} & \tiny{30.54} & \tiny{63.70} & \tiny{62.28} & \tiny{68.6} & \tiny{52.8} & \tiny{64.54} & \tiny{50.13} & \tiny{57.34} \\
    \tiny{No} & \tiny{100\%} & \tiny{62.71} & \tiny{45.98} & \tiny{77.98} & \tiny{58.83} & \tiny{52.07} & \tiny{33.01} & \tiny{65.38} & \tiny{62.09} & \tiny{70.4} & \tiny{54.3} & \tiny{65.71} & \tiny{50.84} & \tiny{58.28} \\
\noalign{\hrule height 2pt}
    \tiny{Yes} & \tiny{10\%} & \tiny{69.45} & \tiny{48.37} & \tiny{84.34} & \tiny{62.03} & \tiny{55.20} & \tiny{33.56} & \tiny{72.75} & \tiny{72.52} & \tiny{72.0} & \tiny{53.1} & \tiny{70.75} & \tiny{53.92} & \tiny{62.33} \\
    \tiny{Yes} & \tiny{20\%} & \tiny{68.08} & \tiny{47.01} & \tiny{84.13} & \tiny{61.32} & \tiny{54.30} & \tiny{33.34} & \tiny{70.15} & \tiny{69.65} & \tiny{72.3} & \tiny{52.8} & \tiny{69.79} & \tiny{52.82} & \tiny{61.31} \\
    \tiny{Yes} & \tiny{30\%} & \tiny{67.91} & \tiny{47.52} & \tiny{84.13} & \tiny{62.28} & \tiny{54.46} & \tiny{34.80} & \tiny{72.47} & \tiny{73.17} & \tiny{71.8} & \tiny{55.5} & \tiny{70.15} & \tiny{54.65} & \tiny{62.40} \\
    \tiny{Yes} & \tiny{40\%} & \tiny{68.08} & \tiny{47.44} & \tiny{83.58} & \tiny{62.41} & \tiny{53.88} & \tiny{33.69} & \tiny{70.36} & \tiny{71.67} & \tiny{72.6} & \tiny{53.8} & \tiny{69.70} & \tiny{53.80} & \tiny{61.75} \\
    \tiny{Yes} & \tiny{50\%} & \tiny{69.11} & \tiny{48.38} & \tiny{83.88} & \tiny{63.26} & \tiny{54.00} & \tiny{34.05} & \tiny{73.58} & \tiny{74.30} & \tiny{71.1} & \tiny{54.0} & \tiny{70.33} & \tiny{54.80} & \tiny{62.57} \\
    \tiny{Yes} & \tiny{60\%} & \tiny{67.15} & \tiny{47.86} & \tiny{83.37} & \tiny{62.92} & \tiny{53.61} & \tiny{33.34} & \tiny{75.16} & \tiny{75.55} & \tiny{70.9} & \tiny{53.0} & \tiny{70.04} & \tiny{54.53} & \tiny{62.28} \\
    \tiny{Yes} & \tiny{70\%} & \tiny{67.15} & \tiny{47.95} & \tiny{83.16} & \tiny{62.75} & \tiny{53.55} & \tiny{34.17} & \tiny{73.57} & \tiny{72.77} & \tiny{71.6} & \tiny{54.3} & \tiny{69.80} & \tiny{54.39} & \tiny{62.10} \\
    \tiny{Yes} & \tiny{80\%} & \tiny{67.58} & \tiny{46.08} & \tiny{82.95} & \tiny{62.54} & \tiny{51.69} & \tiny{32.10} & \tiny{73.12} & \tiny{73.66} & \tiny{70.0} & \tiny{51.7} & \tiny{69.06} & \tiny{53.21} & \tiny{61.14} \\
    \tiny{Yes} & \tiny{90\%} & \tiny{63.91} & \tiny{47.18} & \tiny{79.88} & \tiny{60.35} & \tiny{48.89} & \tiny{31.31} & \tiny{69.51} & \tiny{62.96} & \tiny{68.7} & \tiny{54.0} & \tiny{66.18} & \tiny{51.16} & \tiny{58.70} \\
    \tiny{Yes} & \tiny{100\%} & \tiny{68.00} & \tiny{48.63} & \tiny{83.12} & \tiny{62.96} & \tiny{52.87} & \tiny{35.91} & \tiny{70.06} & \tiny{67.85} & \tiny{71.8} & \tiny{55.8} & \tiny{69.17} & \tiny{54.23} & \tiny{61.70} \\
\noalign{\hrule height 2pt}
    \multicolumn{2}{|c|}{\tiny{Original}} & \tiny{62.12} & \tiny{40.70} & \tiny{74.12} & \tiny{52.48} & \tiny{50.37} & \tiny{31.30} & \tiny{62.72} & \tiny{62.22} & \tiny{68.6} & \tiny{41.2} & \tiny{63.58} & \tiny{45.58} & \tiny{54.58} \\
\noalign{\hrule height 2pt}
\end{tabular}
\caption{Results (.2f) from each training attempt with 5\% of our training data over Llama 3.2 3B }
\label{table:102}
\end{table*}
\begin{table*}[!h]
\centering
\begin{tabular}{cc|cc|cc|cc|cc|cc|ccc}
\noalign{\hrule height 2pt}
\rowcolor{green} \textbf{\tiny{Benchmarks}}& \textbf{\tiny{Ratio of}} & \multicolumn{2}{|c|}{\tiny{\textbf{ARC-Challenge}}} & \multicolumn{2}{|c|}{\tiny{\textbf{ARC-Easy}}} & \multicolumn{2}{|c|}{\tiny{\textbf{MMLU}}} & \multicolumn{2}{|c|}{\tiny{\textbf{BoolQ}}} & \multicolumn{2}{|c|}{\tiny{\textbf{Context-MCQ}}} & \multicolumn{3}{|c|}{\tiny{\textbf{Overall Average}}} \\
\cline{2-15}
\rowcolor{green} \textbf{\tiny{Data used?}}& \textbf{\tiny{Hindi}} & \textbf{\tiny{En}} & \textbf{\tiny{Hi}} & \textbf{\tiny{En}} & \textbf{\tiny{Hi}} & \textbf{\tiny{En}} & \textbf{\tiny{Hi}} & \textbf{\tiny{En}} & \textbf{\tiny{Hi}} & \textbf{\tiny{En}} & \textbf{\tiny{Hi}} & \textbf{\tiny{En}} & \textbf{\tiny{Hi}} & \textbf{\tiny{Tot}} \\
\noalign{\hrule height 2pt}
\tiny{No} & \tiny{10\%} & \tiny{86.52} & \tiny{75.25} & \tiny{94.52} & \tiny{87.24} & \tiny{68.53} & \tiny{53.93} & \tiny{86.82} & \tiny{83.69} & \tiny{86.7} & \tiny{79.0} & \tiny{84.62} & \tiny{75.82} & \tiny{80.22} \\
\tiny{No} & \tiny{20\%} & \tiny{87.11} & \tiny{75.68} & \tiny{94.57} & \tiny{87.11} & \tiny{68.46} & \tiny{53.89} & \tiny{86.66} & \tiny{83.42} & \tiny{86.9} & \tiny{78.6} & \tiny{84.74} & \tiny{75.80} & \tiny{80.27} \\
\tiny{No} & \tiny{30\%} & \tiny{86.34} & \tiny{75.42} & \tiny{94.86} & \tiny{87.28} & \tiny{68.74} & \tiny{53.85} & \tiny{86.91} & \tiny{83.94} & \tiny{87.2} & \tiny{78.4} & \tiny{84.81} & \tiny{75.42} & \tiny{80.29} \\
\tiny{No} & \tiny{40\%} & \tiny{86.86} & \tiny{75.85} & \tiny{95.32} & \tiny{87.45} & \tiny{68.88} & \tiny{54.36} & \tiny{86.60} & \tiny{83.76} & \tiny{86.8} & \tiny{78.1} & \tiny{84.89} & \tiny{75.91} & \tiny{80.40} \\
\tiny{No} & \tiny{50\%} & \tiny{86.86} & \tiny{75.51} & \tiny{95.11} & \tiny{87.41} & \tiny{68.49} & \tiny{53.96} & \tiny{86.82} & \tiny{84.06} & \tiny{87.1} & \tiny{77.8} & \tiny{84.88} & \tiny{75.75} & \tiny{80.31} \\
\tiny{No} & \tiny{60\%} & \tiny{87.11} & \tiny{76.62} & \tiny{95.70} & \tiny{87.83} & \tiny{68.43} & \tiny{53.73} & \tiny{86.60} & \tiny{84.15} & \tiny{87.2} & \tiny{78.3} & \tiny{85.01} & \tiny{76.12} & \tiny{80.57} \\
\tiny{No} & \tiny{70\%} & \tiny{88.65} & \tiny{78.07} & \tiny{95.16} & \tiny{89.27} & \tiny{71.32} & \tiny{56.13} & \tiny{87.76} & \tiny{85.01} & \tiny{88.3} & \tiny{79.1} & \tiny{86.24} & \tiny{77.51} & \tiny{81.88} \\
\tiny{No} & \tiny{80\%} & \tiny{88.22} & \tiny{77.47} & \tiny{95.24} & \tiny{88.93} & \tiny{70.00} & \tiny{55.06} & \tiny{87.19} & \tiny{85.13} & \tiny{87.1} & \tiny{85.13} & \tiny{85.55} & \tiny{77.04} & \tiny{81.30} \\
\tiny{No} & \tiny{90\%} & \tiny{86.94} & \tiny{76.00} & \tiny{95.28} & \tiny{87.58} & \tiny{69.42} & \tiny{54.61} & \tiny{86.48} & \tiny{84.12} & \tiny{87.0} & \tiny{79.2} & \tiny{85.02} & \tiny{76.30} & \tiny{80.66} \\
\tiny{No} & \tiny{100\%} & \tiny{88.48} & \tiny{76.36} & \tiny{95.37} & \tiny{89.10} & \tiny{70.00} & \tiny{54.36} & \tiny{86.64} & \tiny{84.34} & \tiny{87.1} & \tiny{79.1} & \tiny{85.52} & \tiny{76.65} & \tiny{81.08} \\
\noalign{\hrule height 2pt}
\tiny{Yes} & \tiny{10\%} & \tiny{87.79} & \tiny{78.24} & \tiny{95.70} & \tiny{90.27} & \tiny{68.87} & \tiny{54.18} & \tiny{86.85} & \tiny{84.91} & \tiny{87.2} & \tiny{79.1} & \tiny{85.28} & \tiny{77.34} & \tiny{81.31} \\
\tiny{Yes} & \tiny{20\%} & \tiny{87.54} & \tiny{77.81} & \tiny{95.45} & \tiny{90.31} & \tiny{68.76} & \tiny{53.99} & \tiny{86.85} & \tiny{84.91} & \tiny{87.5} & \tiny{79.8} & \tiny{85.22} & \tiny{77.36} & \tiny{81.29} \\
\tiny{Yes} & \tiny{30\%} & \tiny{87.88} & \tiny{78.41} & \tiny{95.87} & \tiny{90.10} & \tiny{68.87} & \tiny{54.60} & \tiny{86.81} & \tiny{85.19} & \tiny{87.4} & \tiny{79.3} & \tiny{85.37} & \tiny{77.50} & \tiny{81.44} \\
\tiny{Yes} & \tiny{40\%} & \tiny{87.80} & \tiny{77.38} & \tiny{94.91} & \tiny{89.86} & \tiny{68.25} & \tiny{53.56} & \tiny{86.85} & \tiny{84.83} & \tiny{87.5} & \tiny{79.3} & \tiny{85.06} & \tiny{77.39} & \tiny{81.02} \\
\tiny{Yes} & \tiny{50\%} & \tiny{87.46} & \tiny{77.73} & \tiny{95.37} & \tiny{90.28} & \tiny{68.25} & \tiny{53.57} & \tiny{86.97} & \tiny{84.89} & \tiny{87.2} & \tiny{79.7} & \tiny{85.05} & \tiny{77.23} & \tiny{81.14} \\
\tiny{Yes} & \tiny{60\%} & \tiny{88.31} & \tiny{78.41} & \tiny{95.74} & \tiny{90.65} & \tiny{68.62} & \tiny{54.18} & \tiny{86.81} & \tiny{85.19} & \tiny{88.0} & \tiny{78.9} & \tiny{85.50} & \tiny{77.47} & \tiny{81.48} \\
\tiny{Yes} & \tiny{70\%} & \tiny{89.16} & \tiny{78.84} & \tiny{95.20} & \tiny{89.56} & \tiny{71.17} & \tiny{56.20} & \tiny{88.04} & \tiny{85.56} & \tiny{88.5} & \tiny{78.4} & \tiny{86.42} & \tiny{77.71} & \tiny{82.06} \\
\tiny{Yes} & \tiny{80\%} & \tiny{87.62} & \tiny{78.58} & \tiny{95.45} & \tiny{89.94} & \tiny{67.91} & \tiny{52.55} & \tiny{86.88} & \tiny{84.12} & \tiny{87.6} & \tiny{78.1} & \tiny{85.09} & \tiny{76.66} & \tiny{80.87} \\
\tiny{Yes} & \tiny{90\%} & \tiny{88.22} & \tiny{78.66} & \tiny{95.37} & \tiny{90.19} & \tiny{68.59} & \tiny{53.70} & \tiny{86.85} & \tiny{84.30} & \tiny{87.5} & \tiny{79.8} & \tiny{85.30} & \tiny{77.33} & \tiny{81.32} \\
\tiny{Yes} & \tiny{100\%} & \tiny{87.88} & \tiny{78.24} & \tiny{95.03} & \tiny{90.02} & \tiny{69.21} & \tiny{53.31} & \tiny{87.00} & \tiny{85.44} & \tiny{87.7} & \tiny{79.4} & \tiny{85.37} & \tiny{77.28} & \tiny{81.32} \\
\noalign{\hrule height 2pt}
\multicolumn{2}{|c|}{\tiny{Original}} & \tiny{88.74} & \tiny{79.18} & \tiny{95.33} & \tiny{88.76} & \tiny{71.00} & \tiny{56.14} & \tiny{87.89} & \tiny{84.67} & \tiny{88.2} & \tiny{77.3} & \tiny{86.23} & \tiny{77.21} & \tiny{81.72} \\
\noalign{\hrule height 2pt}
\end{tabular}
\caption{Results (.2f) from each training attempt with 5\% of our training data over Gemma 2 9B }
 \label{table:103}
\end{table*}
\begin{table*}[!h]
\centering
\begin{tabular}{cc|cc|cc|cc|cc|cc|ccc}
 \noalign{\hrule height 2pt}
 \rowcolor{green}   \textbf{\tiny{Benchmarks}}& \textbf{\tiny{Ratio of}} & \multicolumn{2}{|c|}{\tiny{\textbf{ARC-Challenge}}} & \multicolumn{2}{|c|}{\tiny{\textbf{ARC-Easy}}} & \multicolumn{2}{|c|}{\tiny{\textbf{MMLU}}} & \multicolumn{2}{|c|}{\tiny{\textbf{BoolQ}}} & \multicolumn{2}{|c|}{\tiny{\textbf{Context-MCQ}}} & \multicolumn{3}{|c|}{\tiny{\textbf{Overall Average}}} \\
 \cline{2-15}
 \rowcolor{green}   \textbf{\tiny{Data used?}}& \textbf{\tiny{Hindi}} & \textbf{\tiny{En}} & \textbf{\tiny{Hi}} & \textbf{\tiny{En}} & \textbf{\tiny{Hi}} & \textbf{\tiny{En}} & \textbf{\tiny{Hi}} & \textbf{\tiny{En}} & \textbf{\tiny{Hi}} & \textbf{\tiny{En}} & \textbf{\tiny{Hi}} & \textbf{\tiny{En}} & \textbf{\tiny{Hi}} & \textbf{\tiny{Tot}} \\
\noalign{\hrule height 2pt}
    \tiny{No} & \tiny{10\%} & \tiny{65.36} & \tiny{45.39} & \tiny{80.26} & \tiny{58.96} & \tiny{49.54} & \tiny{35.22} & \tiny{77.22} & \tiny{75.19} & \tiny{64.7} & \tiny{54.6} & \tiny{67.42} & \tiny{53.87} & \tiny{60.64} \\
    \tiny{No} & \tiny{20\%} & \tiny{64.93} & \tiny{45.31} & \tiny{80.01} & \tiny{58.80} & \tiny{49.20} & \tiny{35.08} & \tiny{76.64} & \tiny{74.89} & \tiny{64.4} & \tiny{54.0} & \tiny{67.04} & \tiny{53.61} & \tiny{60.32} \\
    \tiny{No} & \tiny{30\%} & \tiny{64.68} & \tiny{46.67} & \tiny{80.35} & \tiny{59.43} & \tiny{49.53} & \tiny{35.17} & \tiny{76.06} & \tiny{74.92} & \tiny{65.0} & \tiny{54.6} & \tiny{67.12} & \tiny{54.16} & \tiny{60.64} \\
    \tiny{No} & \tiny{40\%} & \tiny{70.22} & \tiny{49.66} & \tiny{83.63} & \tiny{63.97} & \tiny{52.08} & \tiny{36.83} & \tiny{81.83} & \tiny{76.48} & \tiny{68.0} & \tiny{57.6} & \tiny{71.15} & \tiny{56.91} & \tiny{64.03} \\
    \tiny{No} & \tiny{50\%} & \tiny{61.86} & \tiny{45.81} & \tiny{79.04} & \tiny{57.99} & \tiny{48.09} & \tiny{34.49} & \tiny{76.54} & \tiny{75.34} & \tiny{63.7} & \tiny{54.0} & \tiny{65.85} & \tiny{53.52} & \tiny{59.69} \\
    \tiny{No} & \tiny{60\%} & \tiny{61.60} & \tiny{45.56} & \tiny{79.58} & \tiny{58.58} & \tiny{47.99} & \tiny{34.39} & \tiny{75.65} & \tiny{75.71} & \tiny{64.6} & \tiny{54.0} & \tiny{65.88} & \tiny{53.65} & \tiny{59.77} \\
    \tiny{No} & \tiny{70\%} & \tiny{63.22} & \tiny{47.78} & \tiny{63.22} & \tiny{59.42} & \tiny{48.26} & \tiny{34.33} & \tiny{76.97} & \tiny{76.13} & \tiny{62.9} & \tiny{52.9} & \tiny{66.33} & \tiny{54.11} & \tiny{60.22} \\
    \tiny{No} & \tiny{80\%} & \tiny{65.53} & \tiny{46.50} & \tiny{81.73} & \tiny{61.03} & \tiny{50.29} & \tiny{35.40} & \tiny{76.79} & \tiny{75.80} & \tiny{64.6} & \tiny{55.3} & \tiny{67.79} & \tiny{54.81} & \tiny{61.30} \\
    \tiny{No} & \tiny{90\%} & \tiny{65.10} & \tiny{46.59} & \tiny{81.73} & \tiny{60.19} & \tiny{50.14} & \tiny{35.41} & \tiny{76.64} & \tiny{75.01} & \tiny{65.0} & \tiny{54.1} & \tiny{67.72} & \tiny{54.26} & \tiny{60.99} \\
    \tiny{No} & \tiny{100\%} & \tiny{67.92} & \tiny{48.81} & \tiny{82.79} & \tiny{62.33} & \tiny{51.42} & \tiny{36.02} & \tiny{80.24} & \tiny{76.14} & \tiny{67.6} & \tiny{56.9} & \tiny{69.99} & \tiny{56.04} & \tiny{63.01} \\
 \noalign{\hrule height 2pt}
    \tiny{Yes} & \tiny{10\%} & \tiny{66.38} & \tiny{48.12} & \tiny{82.24} & \tiny{62.33} & \tiny{49.00} & \tiny{34.76} & \tiny{75.35} & \tiny{72.56} & \tiny{64.2} & \tiny{54.4} & \tiny{67.43} & \tiny{54.43} & \tiny{60.93} \\
    \tiny{Yes} & \tiny{20\%} & \tiny{66.13} & \tiny{48.89} & \tiny{82.24} & \tiny{62.67} & \tiny{48.85} & \tiny{34.84} & \tiny{74.92} & \tiny{71.86} & \tiny{63.8} & \tiny{53.0} & \tiny{67.19} & \tiny{54.25} & \tiny{60.72} \\
    \tiny{Yes} & \tiny{30\%} & \tiny{65.53} & \tiny{48.46} & \tiny{82.15} & \tiny{62.25} & \tiny{49.11} & \tiny{34.87} & \tiny{73.91} & \tiny{71.03} & \tiny{64.2} & \tiny{53.1} & \tiny{66.98} & \tiny{53.94} & \tiny{60.46} \\
    \tiny{Yes} & \tiny{40\%} & \tiny{67.92} & \tiny{48.04} & \tiny{82.45} & \tiny{62.42} & \tiny{50.67} & \tiny{36.23} & \tiny{77.00} & \tiny{75.19} & \tiny{65.4} & \tiny{55.6} & \tiny{68.69} & \tiny{55.49} & \tiny{62.09} \\
    \tiny{Yes} & \tiny{50\%} & \tiny{68.08} & \tiny{51.02} & \tiny{83.96} & \tiny{64.05} & \tiny{47.99} & \tiny{34.64} & \tiny{76.66} & \tiny{74.30} & \tiny{63.9} & \tiny{54.7} & \tiny{68.12} & \tiny{55.74} & \tiny{61.93} \\
    \tiny{Yes} & \tiny{60\%} & \tiny{68.08} & \tiny{50.34} & \tiny{84.21} & \tiny{64.52} & \tiny{47.76} & \tiny{34.62} & \tiny{72.75} & \tiny{70.32} & \tiny{63.5} & \tiny{53.7} & \tiny{67.26} & \tiny{54.70} & \tiny{60.98} \\
    \tiny{Yes} & \tiny{70\%} & \tiny{68.25} & \tiny{51.45} & \tiny{84.55} & \tiny{64.73} & \tiny{48.31} & \tiny{34.78} & \tiny{75.87} & \tiny{73.35} & \tiny{64.6} & \tiny{54.3} & \tiny{68.31} & \tiny{55.72} & \tiny{62.02} \\
    \tiny{Yes} & \tiny{80\%} & \tiny{66.47} & \tiny{49.83} & \tiny{83.50} & \tiny{63.55} & \tiny{48.70} & \tiny{34.62} & \tiny{73.67} & \tiny{69.90} & \tiny{63.4} & \tiny{53.9} & \tiny{67.15} & \tiny{54.36} & \tiny{60.75} \\
    \tiny{Yes} & \tiny{90\%} & \tiny{67.06} & \tiny{49.74} & \tiny{83.42} & \tiny{63.76} & \tiny{49.44} & \tiny{35.32} & \tiny{73.49} & \tiny{69.50} & \tiny{64.2} & \tiny{53.3} & \tiny{67.52} & \tiny{54.32} & \tiny{60.92} \\
    \tiny{Yes} & \tiny{100\%} & \tiny{67.58} & \tiny{49.40} & \tiny{83.00} & \tiny{63.09} & \tiny{50.93} & \tiny{36.01} & \tiny{75.75} & \tiny{73.72} & \tiny{66.0} & \tiny{54.6} & \tiny{68.65} & \tiny{55.36} & \tiny{62.00} \\
 \noalign{\hrule height 2pt}
    \multicolumn{2}{|c|}{\tiny{Original}} & \tiny{71.50} & \tiny{51.62} & \tiny{84.05} & \tiny{64.31} & \tiny{51.13} & \tiny{36.49} & \tiny{82.69} & \tiny{77.12} & \tiny{70.9} & \tiny{59.2} & \tiny{72.05} & \tiny{57.74} & \tiny{64.90} \\
 \noalign{\hrule height 2pt}
 \end{tabular}
 \caption{Results (.2f) from each training attempt with 5\% of our training data over Gemma 2 2B }
 \label{table:104}
\end{table*}
\section{Model Choices}
\label{sec:modelchoices}
\begin{figure*}[!h]
    \centering
    \includegraphics[width=1\linewidth]{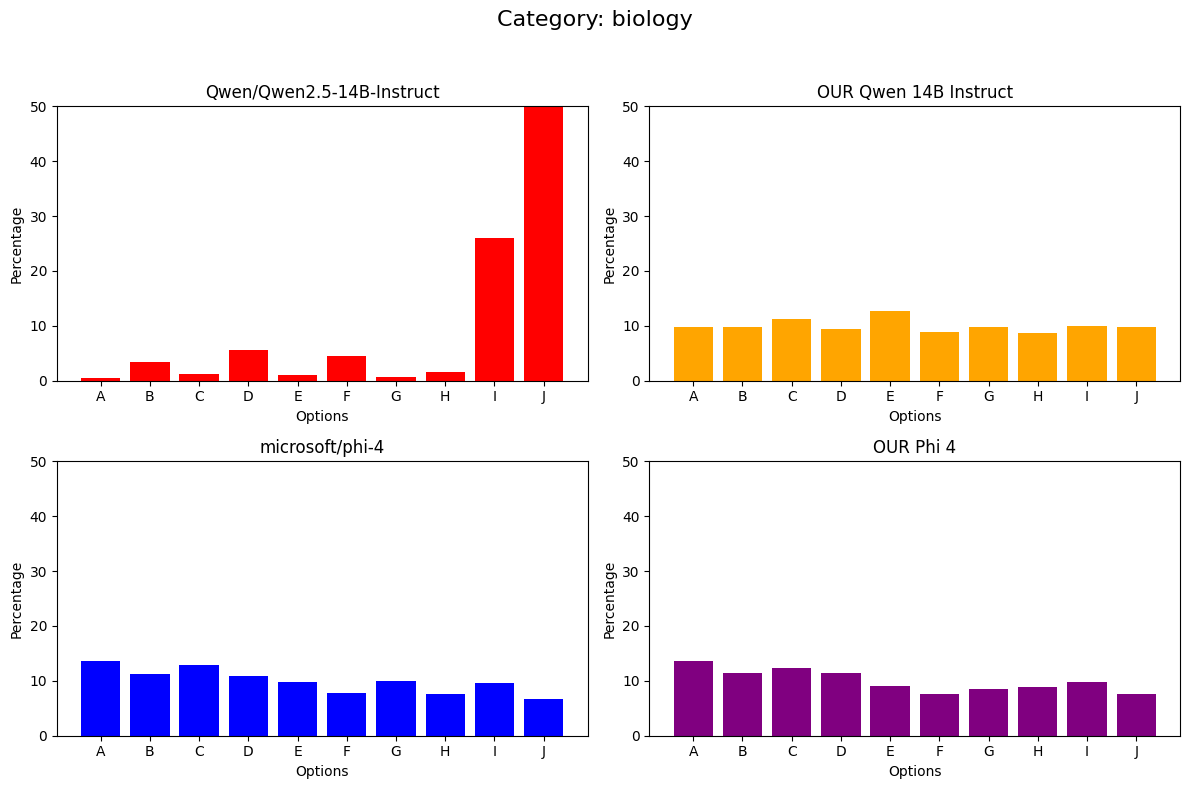}
    \caption{Each model's choice distribution over MMLU-Pro : Biology}
    \label{fig:201}
\end{figure*}

\begin{figure*}[!h]
    \centering
    \includegraphics[width=1\linewidth]{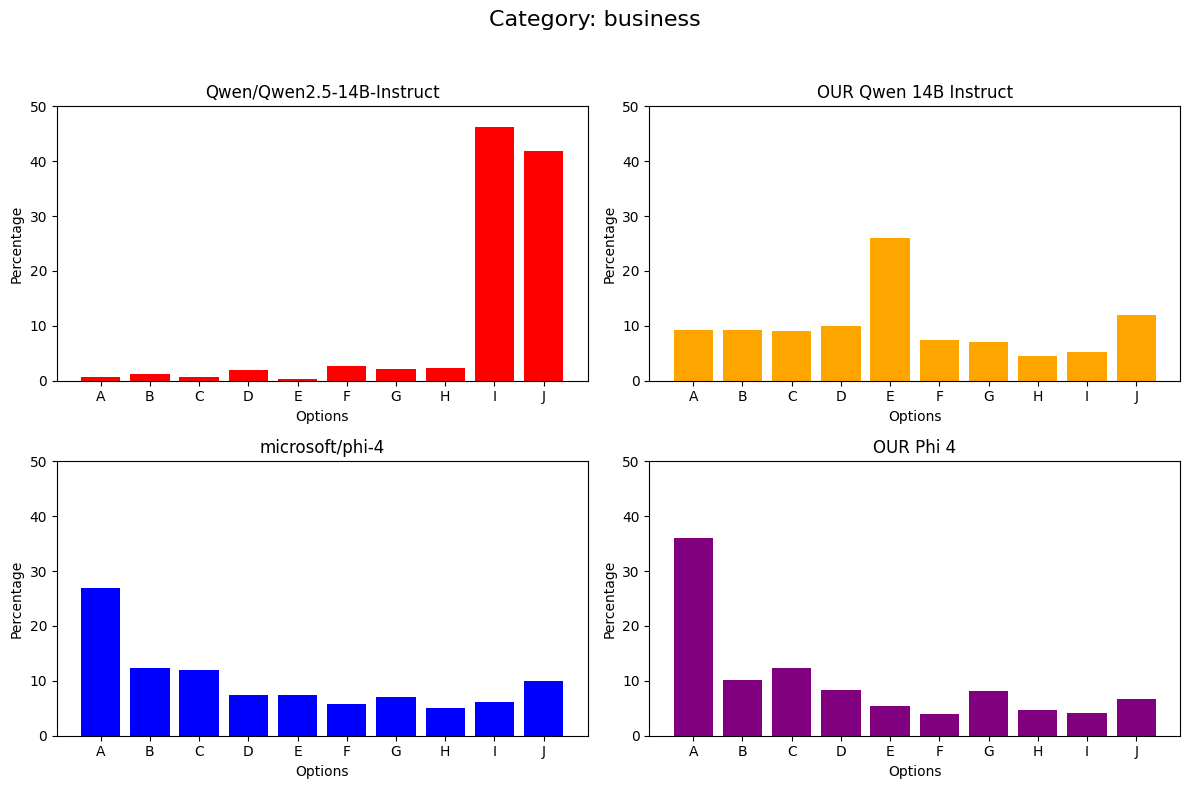}
    \caption{Each model's choice distribution over MMLU-Pro : Business}
    \label{fig:202}
\end{figure*}

\begin{figure*}[!h]
    \centering
    \includegraphics[width=1\linewidth]{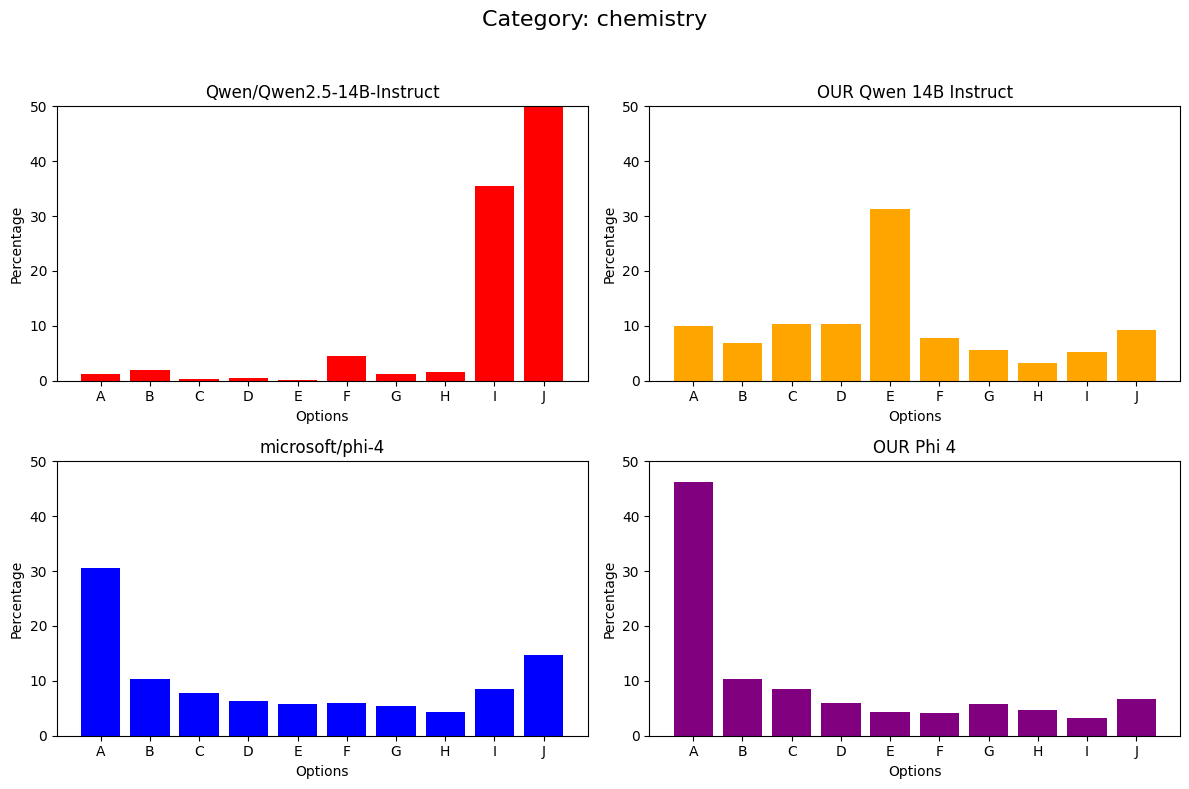}
    \caption{Each model's choice distribution over MMLU-Pro : Chemistry}
    \label{fig:203}
\end{figure*}

\begin{figure*}[!h]
    \centering
    \includegraphics[width=1\linewidth]{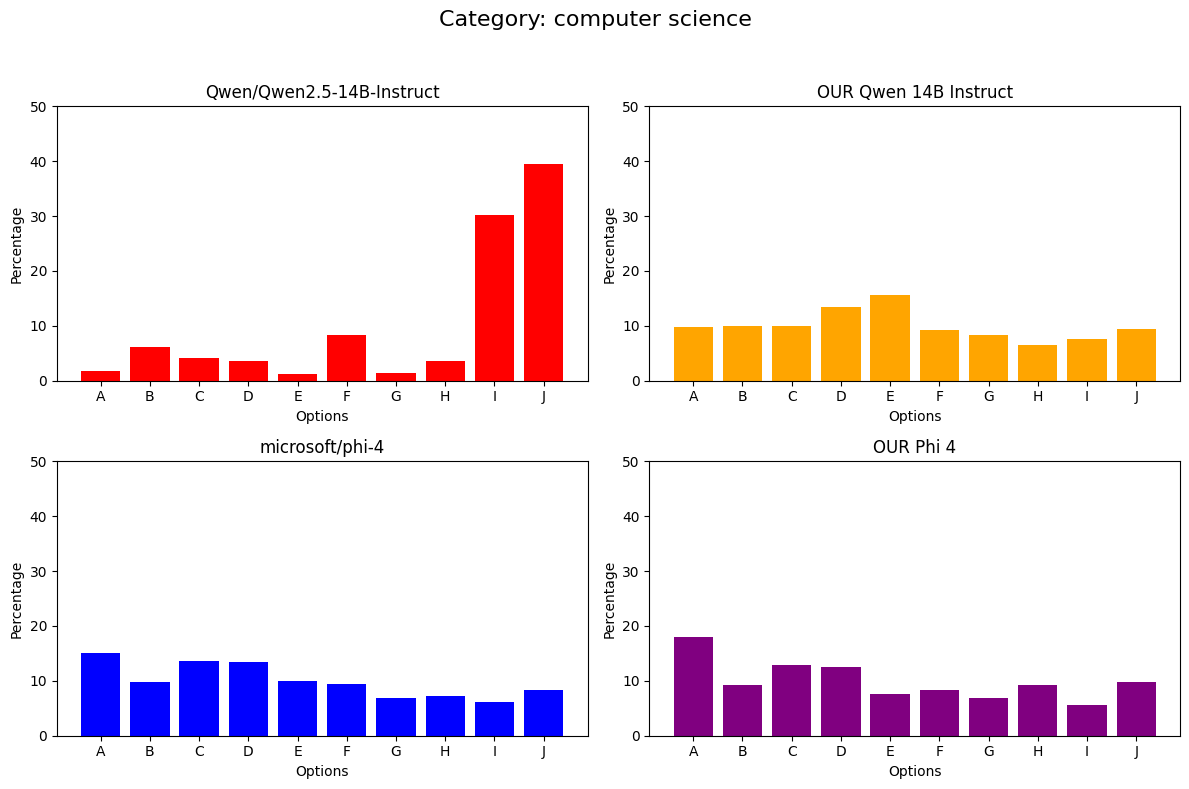}
    \caption{Each model's choice distribution over MMLU-Pro : CS}
    \label{fig:204}
\end{figure*}

\begin{figure*}[!h]
    \centering
    \includegraphics[width=1\linewidth]{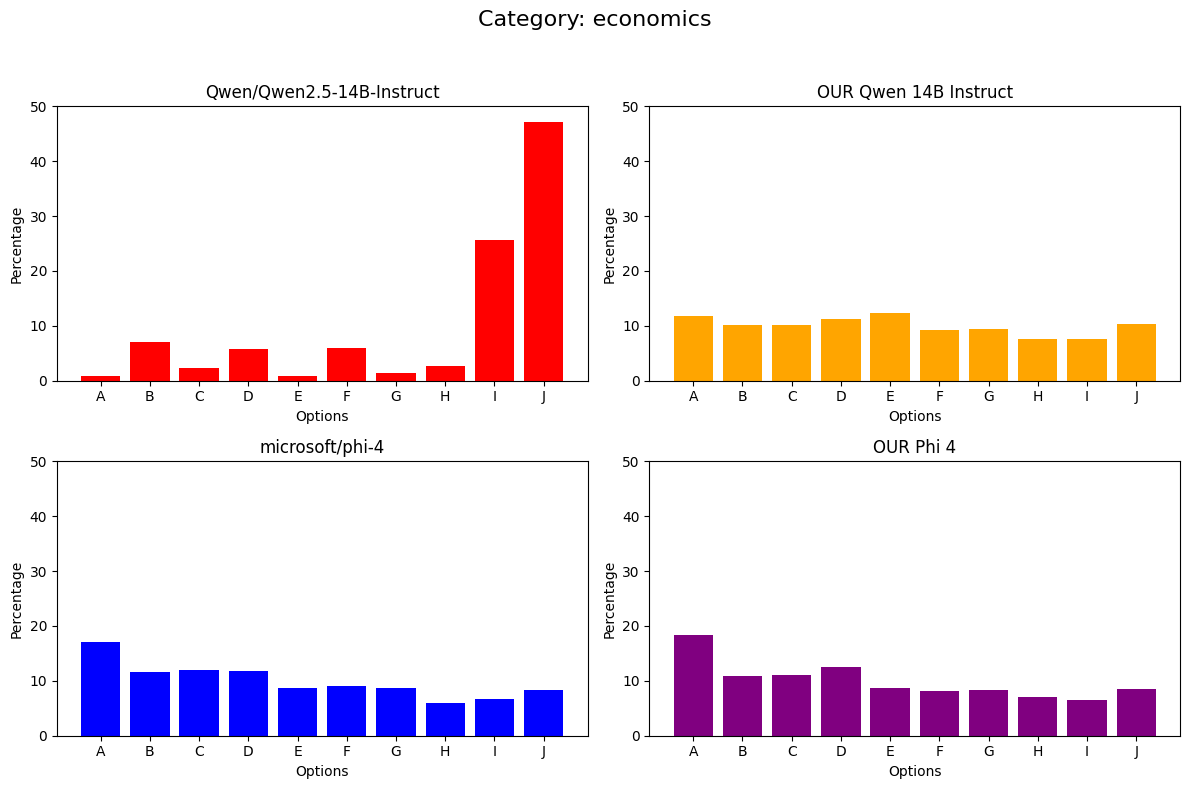}
    \caption{Each model's choice distribution over MMLU-Pro : Economics}
    \label{fig:205}
\end{figure*}

\begin{figure*}[!h]
    \centering
    \includegraphics[width=1\linewidth]{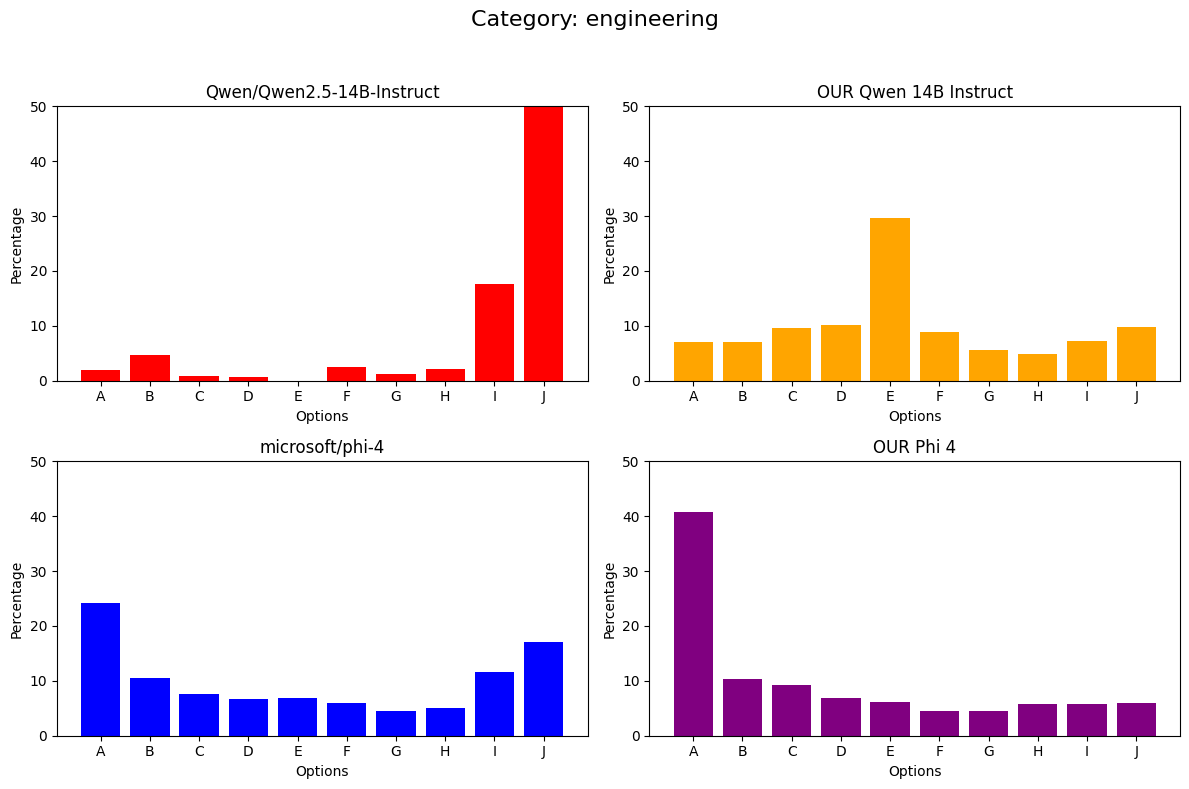}
    \caption{Each model's choice distribution over MMLU-Pro : Engineering}
    \label{fig:206}
\end{figure*}

\begin{figure*}[!h]
    \centering
    \includegraphics[width=1\linewidth]{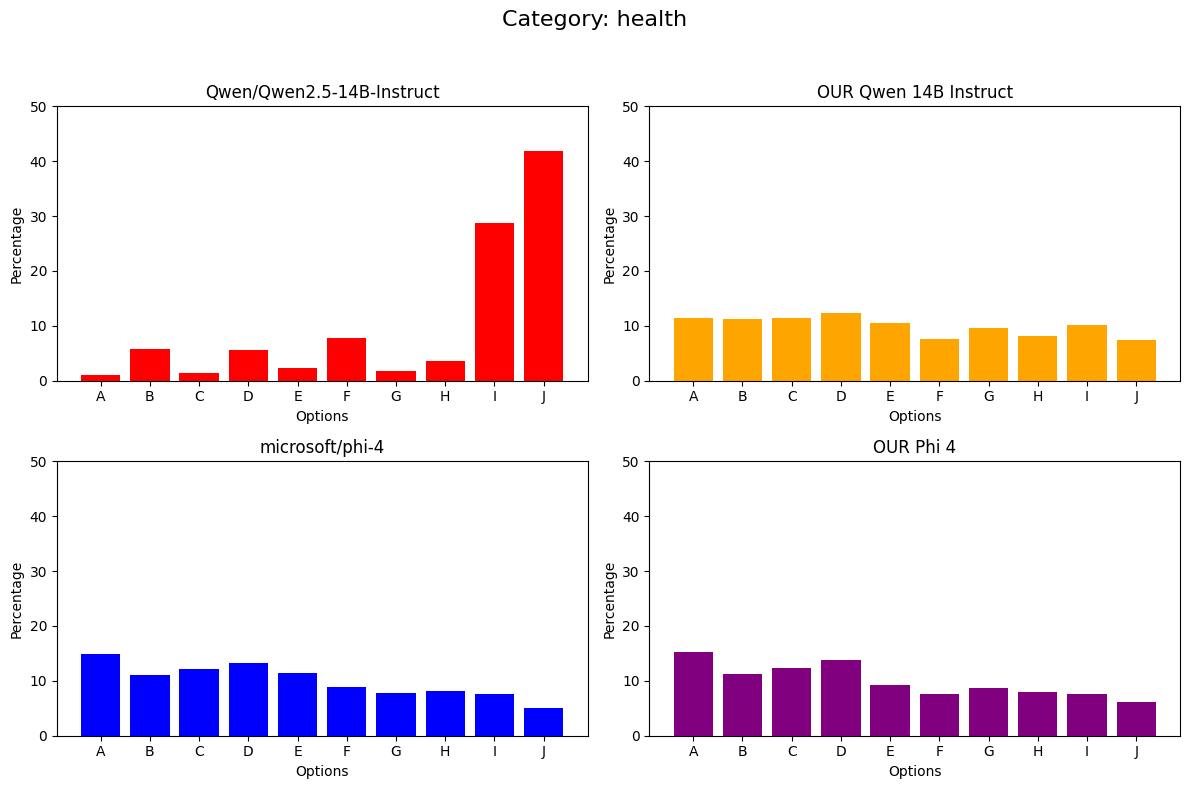}
    \caption{Each model's choice distribution over MMLU-Pro : Health}
    \label{fig:207}
\end{figure*}

\begin{figure*}[!h]
    \centering
    \includegraphics[width=1\linewidth]{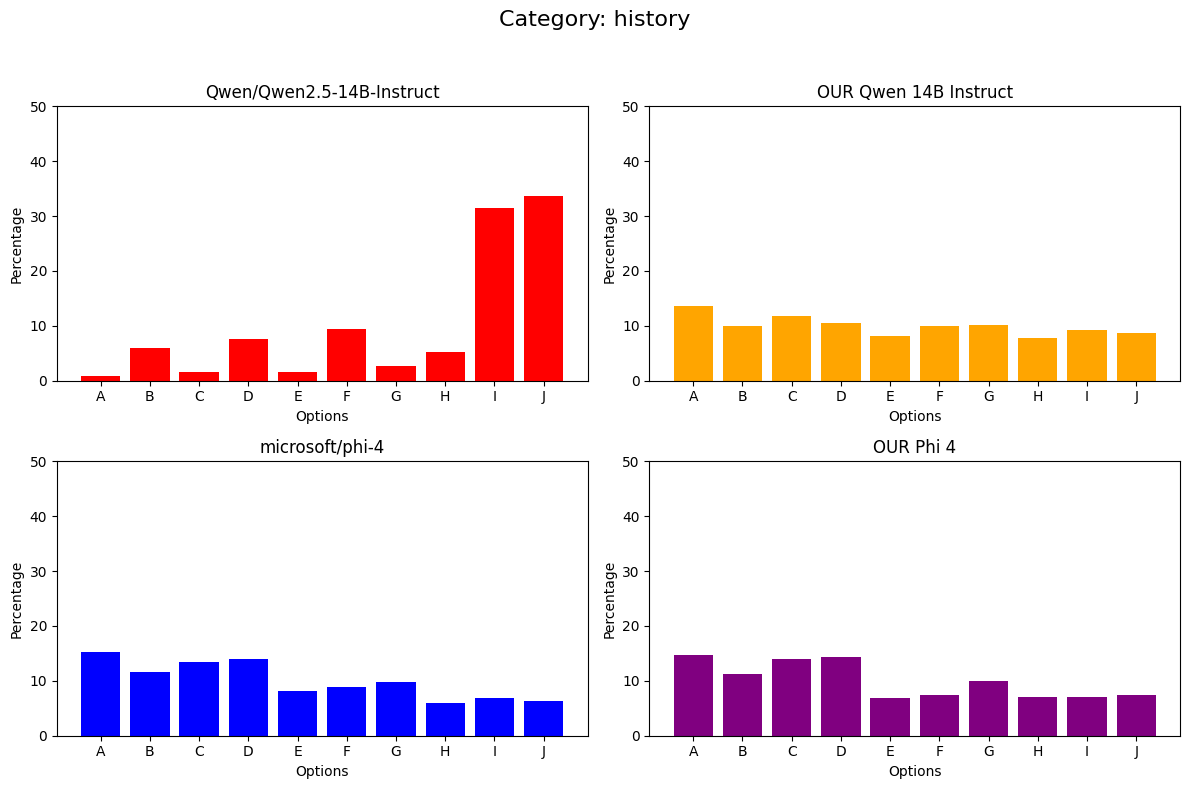}
    \caption{Each model's choice distribution over MMLU-Pro : History}
    \label{fig:208}
\end{figure*}

\begin{figure*}[!h]
    \centering
    \includegraphics[width=1\linewidth]{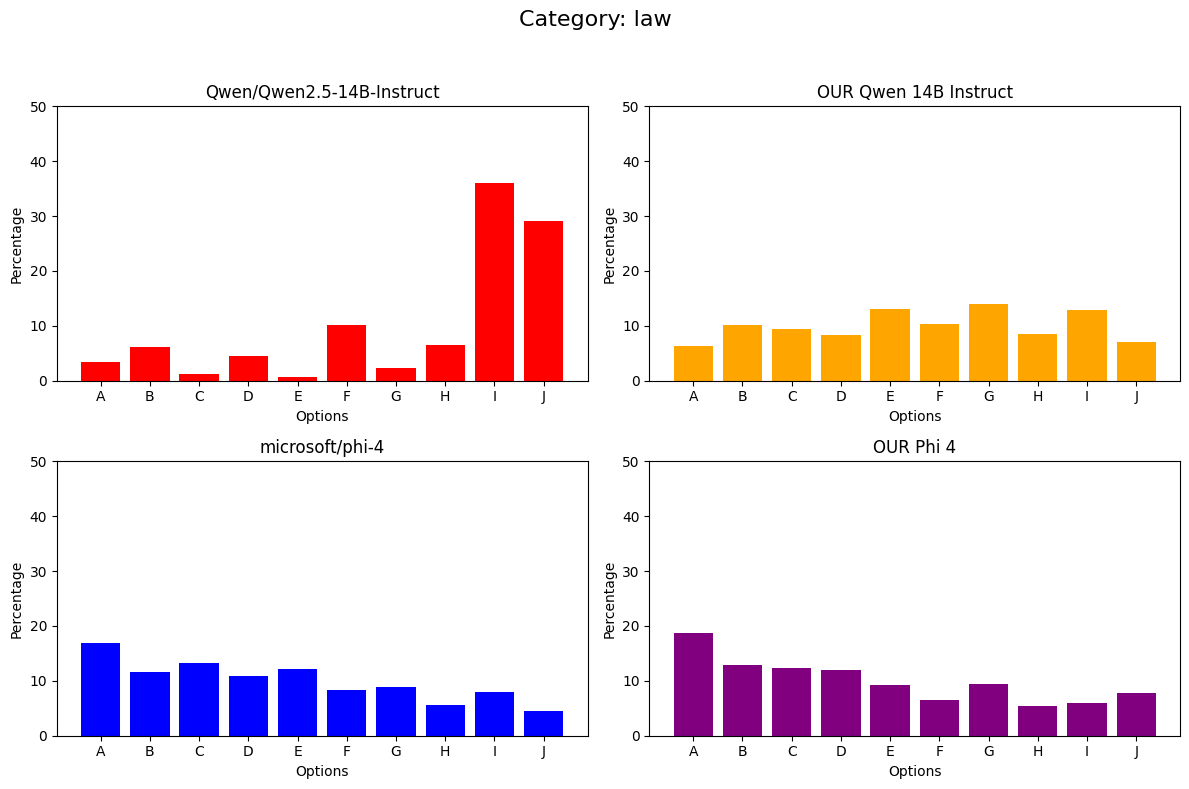}
    \caption{Each model's choice distribution over MMLU-Pro : Law}
    \label{fig:209}
\end{figure*}

\begin{figure*}[!h]
    \centering
    \includegraphics[width=1\linewidth]{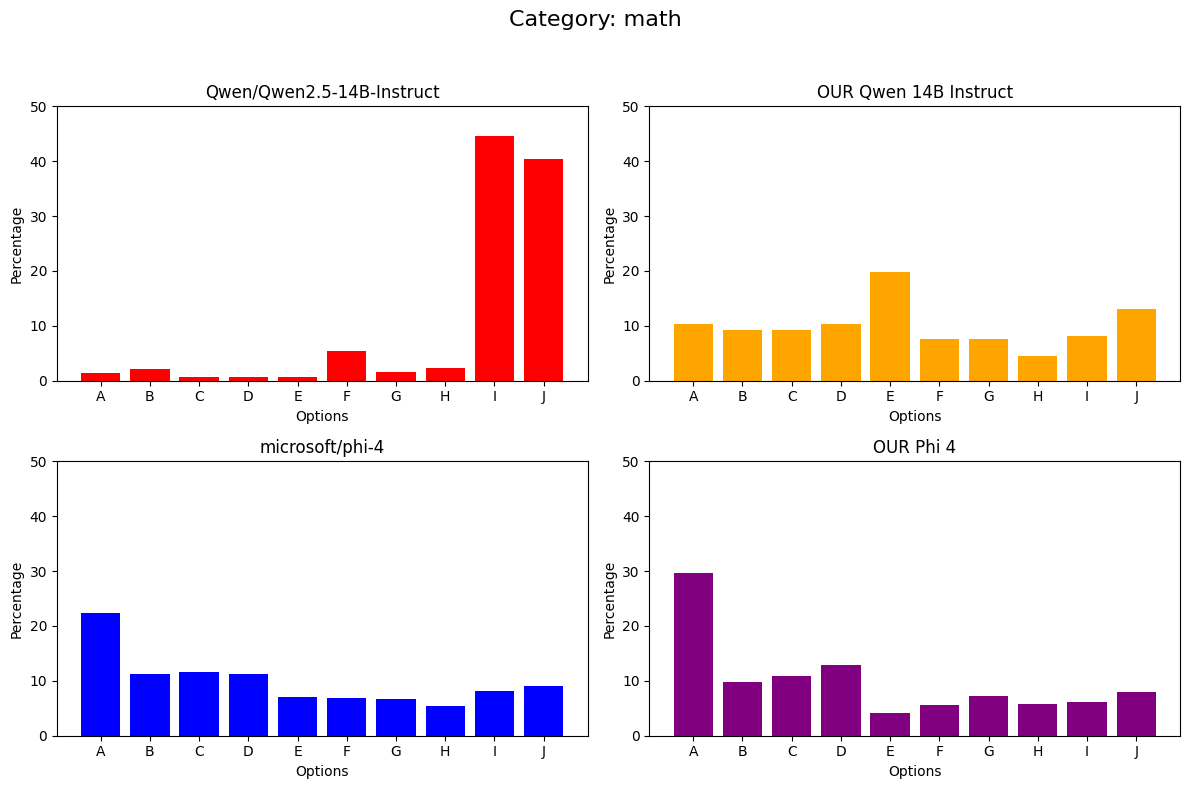}
    \caption{Each model's choice distribution over MMLU-Pro : Math}
    \label{fig:210}
\end{figure*}

\begin{figure*}[!h]
    \centering
    \includegraphics[width=1\linewidth]{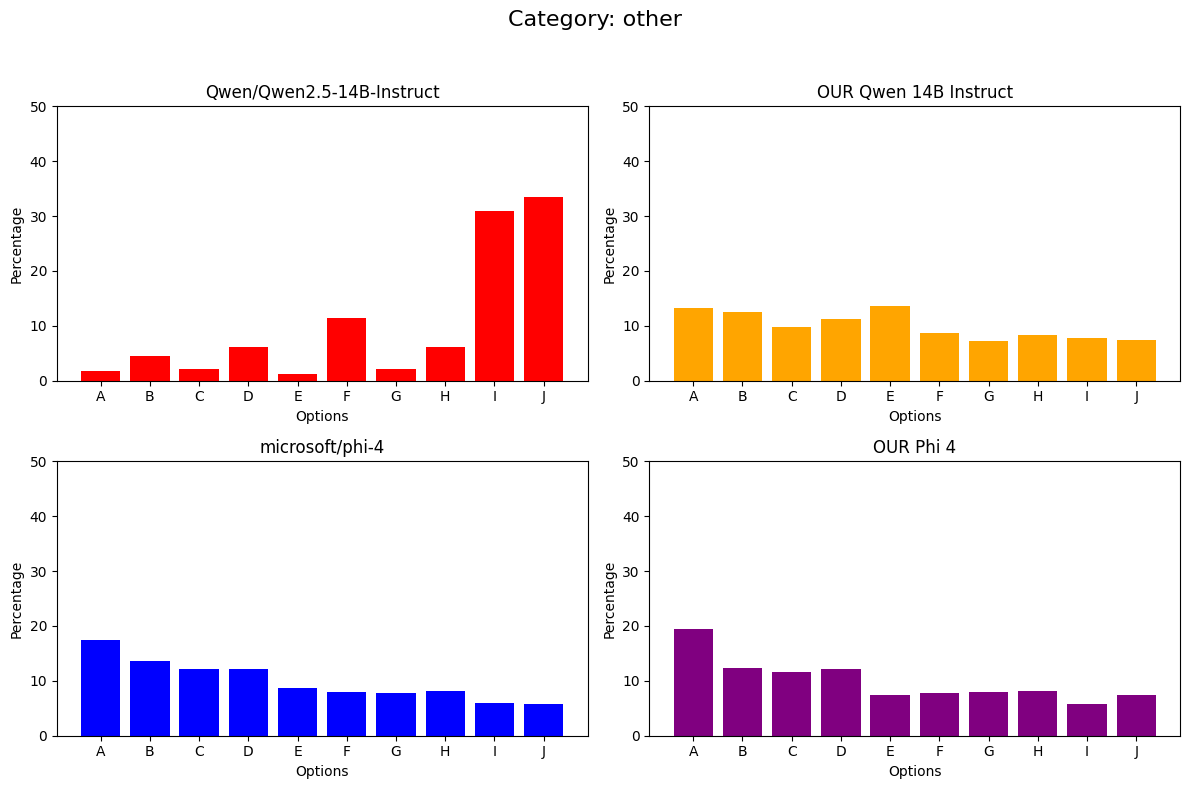}
    \caption{Each model's choice distribution over MMLU-Pro : Other}
    \label{fig:211}
\end{figure*}

\begin{figure*}[!h]
    \centering
    \includegraphics[width=1\linewidth]{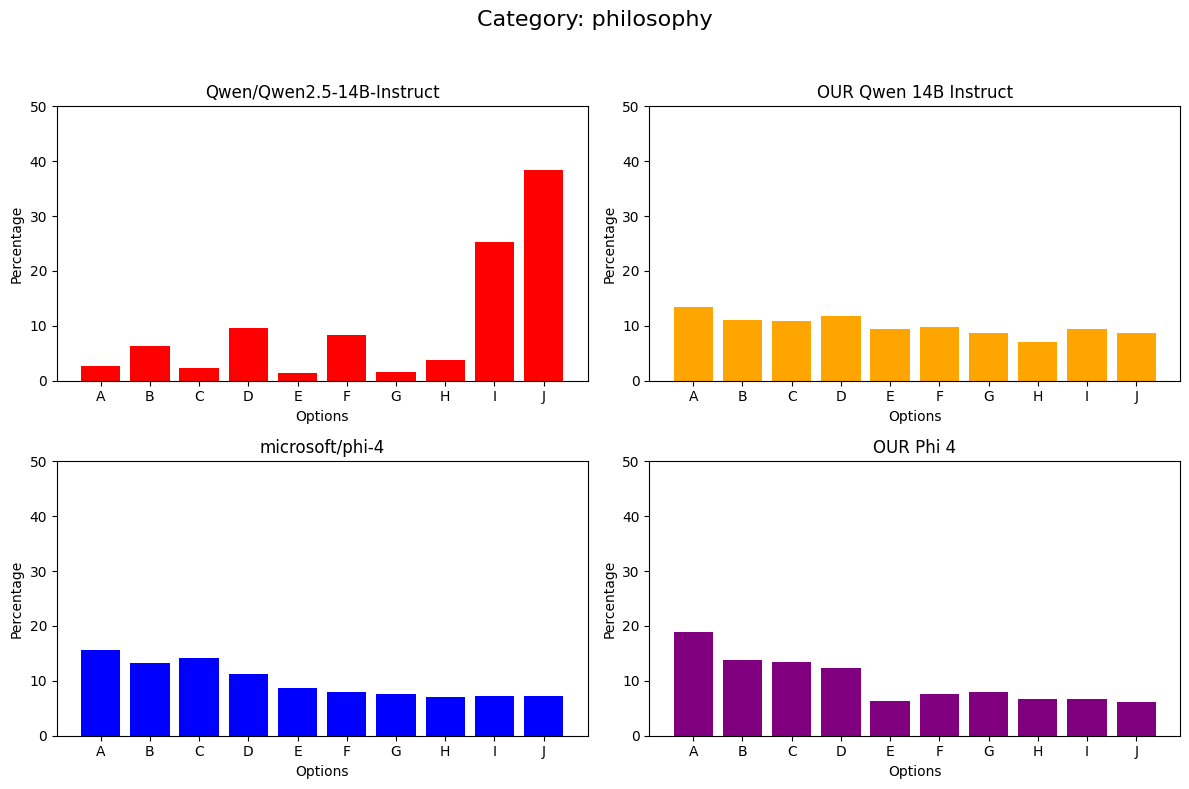}
    \caption{Each model's choice distribution over MMLU-Pro : Philosophy}
    \label{fig:212}
\end{figure*}

\begin{figure*}[!h]
    \centering
    \includegraphics[width=1\linewidth]{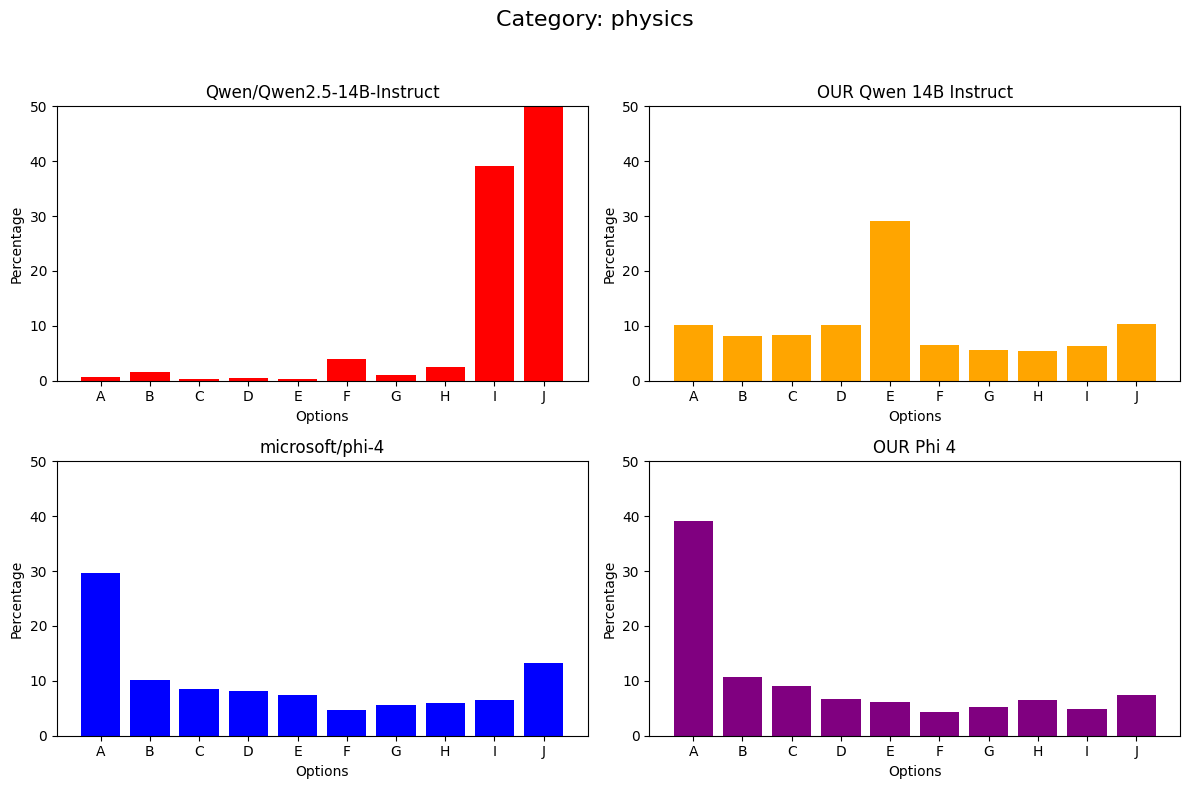}
    \caption{Each model's choice distribution over MMLU-Pro : Physics}
    \label{fig:213}
\end{figure*}

\begin{figure*}[!h]
    \centering
    \includegraphics[width=1\linewidth]{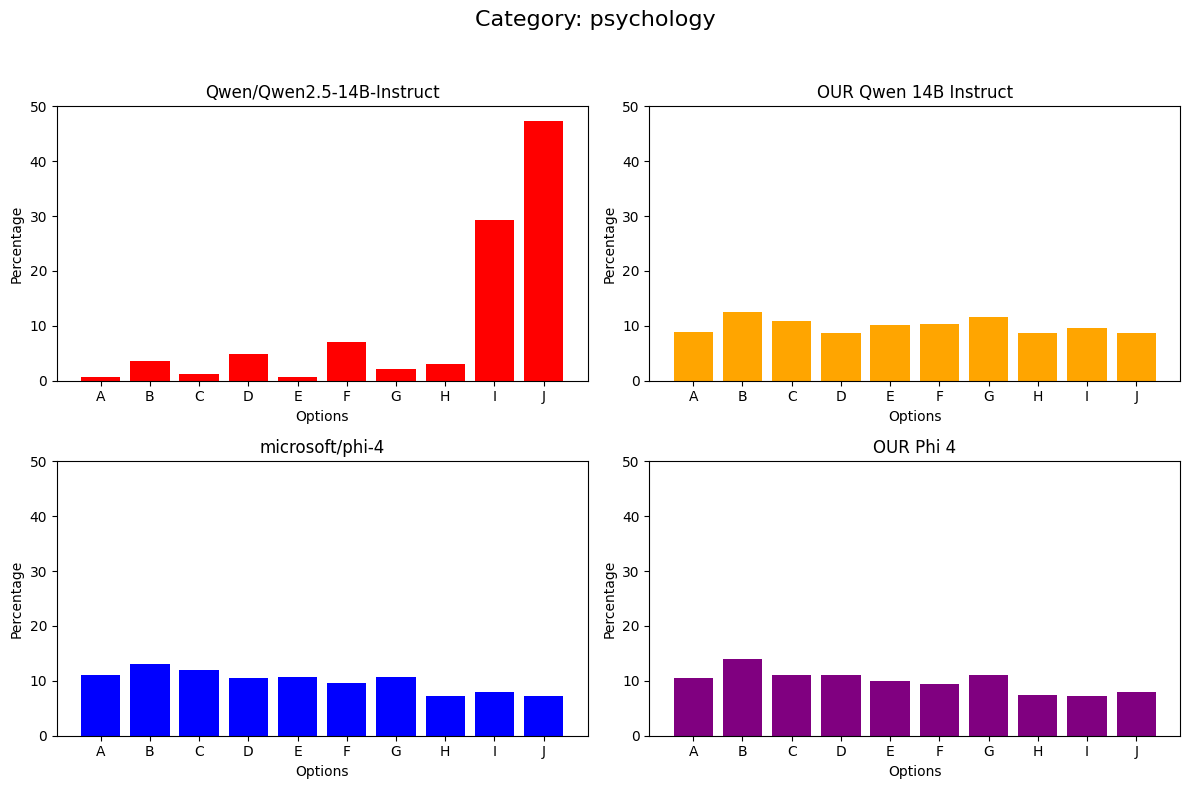}
    \caption{Each model's choice distribution over MMLU-Pro : Psychology}
    \label{fig:214}
\end{figure*}

The choices selected by each of the models over each domain of MMLU-Pro can be seen in the below images \autoref{fig:201} to \autoref{fig:214}.
\end{document}